\documentclass{article}

\usepackage[in]{fullpage}  
\usepackage{natbib} 
\usepackage{mathpazo} 

\usepackage[utf8]{inputenc} %
\usepackage[T1]{fontenc}    %
\usepackage{hyperref}       %
\usepackage{url}            %
\usepackage{booktabs}       %
\usepackage{amsfonts}       %
\usepackage{nicefrac}       %
\usepackage{microtype}      %
\usepackage{lipsum}         %
\usepackage{todonotes}

\newcommand{\reals}{{\mathbb{R}}}

\newcommand{\Tr}{\mathop{\bf Tr}}

\newcommand{\argmin}{\mathop{\rm argmin}}
\newcommand{\argmax}{\mathop{\rm argmax}}

\newcommand{\dquote}[1]{``#1''}

\usepackage{url}            %
\usepackage{booktabs}       %
\usepackage{multirow}    
\usepackage{amsfonts}       %
\usepackage{nicefrac}       %
\usepackage{microtype}      %
\usepackage{natbib}
\usepackage{enumerate}
\usepackage{hhline}
\usepackage{makecell}
\usepackage{pifont}
\usepackage{xspace}

\usepackage{graphicx} %
\usepackage{caption}
\usepackage{subcaption}
\usepackage{amsmath}
\usepackage{amsthm}
\usepackage{amssymb}
\usepackage{tikz}
\usepackage{xcolor}
\usetikzlibrary{arrows}

\allowdisplaybreaks

\usepackage{mathrsfs}

\usepackage{algorithm}
\usepackage{algorithmic}
\usepackage{hyperref}
\usepackage{bm}

\allowdisplaybreaks

\newcommand{\labs}{\left\vert}
\newcommand{\rabs}{\right\vert}
\newcommand{\lnorm}{\left\Vert}
\newcommand{\rnorm}{\right\Vert}

\newcommand{\tr}{\operatorname{tr}}
\newcommand{\opt}{\mathrm{opt}}

\newcommand{\expect}{\mathbb{E}}

\newcommand{\indict}{\mathbb{I}}

\newtheorem{thm}{Theorem}
\newtheorem{lem}{Lemma}

\newtheorem{prop}{Proposition}
\newtheorem{asmp}{Assumption}
\newtheorem{defn}{Definition}

\newtheorem{rem}{Remark}

\usepackage[capitalize,noabbrev]{cleveref}
\crefname{thm}{Theorem}{Theorems}
\crefname{lem}{Lemma}{Lemmas}
\crefname{cor}{Corollary}{Corollaries}
\crefname{prop}{Proposition}{Propositions}
\crefname{asmp}{Assumption}{Assumptions}
\crefname{defn}{Definition}{Definitions}
\crefname{oracle}{Oracle}{Oracles}
\crefname{fact}{Fact}{Facts}
\crefname{conj}{Conjecture}{Conjectures}
\crefname{rem}{Remark}{Remarks}
\crefname{example}{Example}{Examples}
\crefname{condition}{Condition}{Conditions}
\crefname{exercise}{Exercise}{Exercises}
\crefname{algorithm}{Algorithm}{Algorithms}
\crefname{table}{Table}{Tables}
\crefname{figure}{Figure}{Figures}
\crefname{section}{Section}{Sections}
\crefname{subsection}{Section}{Sections}
\crefname{appendix}{Appendix}{Appendices}
\crefname{message}{Message}{Messages}

\newcommand{\algorithmicbreak}{\textbf{break}}
\newcommand{\BREAK}{\STATE \algorithmicbreak}

\definecolor{red}{rgb}{1, 0, 0}

\definecolor{green}{rgb}{0, 1, 0}

\definecolor{blue}{rgb}{0, 0, 1}

\definecolor{orange}{rgb}{1, 0.4, 0.0}

\input{packages/math_commands}

\newcommand{\est}{\operatorname{EST}}
\newcommand{\rfe}{\operatorname{RFE}}
\newcommand{\env}{\operatorname{env}}
\renewcommand{\opt}{\operatorname{OPT}}
\newcommand{\rl}{\operatorname{RL}}
\newcommand{\bc}{\operatorname{BC}}

\usepackage{tabulary}

\usepackage{amssymb}%
\usepackage{pifont}%

\AtBeginDocument{\setlength\abovedisplayskip{5pt}}
\AtBeginDocument{\setlength\belowdisplayskip{5pt}}

\hypersetup{
    colorlinks=true,
    citecolor=blue,
    linkcolor=blue,
}

\newcommand{\expert}{\operatorname{E}}
\newcommand{\piE}{\pi^{\expert}}

\newcommand{\mbalgname}{MB-TAIL\xspace}

\title{Provably Efficient Adversarial Imitation Learning with Unknown Transitions}

\usepackage{authblk}
\author[1, 4]{Tian Xu\thanks{Equal contribution. Author ordering is determined randomly using a coin flip.}}{}
\author[2, 3]{{Ziniu Li{$^*$}}}
\author[1, 4]{{{Yang Yu\thanks{Corresponding author.}}}}
\author[2, 3]{{Zhi-Quan Luo{$^\dag$}}}
\affil[1]{%
    National Key Laboratory for Novel Software Technology, Nanjing University
}
\affil[2]{%
The Chinese University of Hong Kong, Shenzhen
}
\affil[3]{%
Shenzhen Research Institute of Big Data
  }
\affil[4]{%
Polixir.ai
  }

\date{\today}

\begin{document}

\maketitle
\begin{abstract}
Imitation learning (IL) has proven to be an effective method for learning good policies from expert demonstrations. Adversarial imitation learning (AIL), a subset of IL methods, is particularly promising, but its theoretical foundation in the presence of unknown transitions has yet to be fully developed. This paper explores the theoretical underpinnings of AIL in this context, where the stochastic and uncertain nature of environment transitions presents a challenge.  We examine the expert sample complexity and interaction complexity required to recover good policies. To this end, we establish a framework connecting reward-free exploration and AIL, and propose an algorithm, MB-TAIL, that achieves the minimax optimal expert sample complexity of $\widetilde{\mathcal{O}} (H^{3/2} |\mathcal{S}|/\varepsilon)$ and interaction complexity of $\widetilde{\mathcal{O}} (H^{3} |\mathcal{S}|^2 |\mathcal{A}|/\varepsilon^2)$. Here, $H$ represents the planning horizon, $|\gS|$ is the state space size, $|\gA|$ is the action space size, and $\varepsilon$ is the desired imitation gap. MB-TAIL is the first algorithm to achieve this level of expert sample complexity in the unknown transition setting and improves upon the interaction complexity of the best-known algorithm, OAL, by $\gO (H)$. Additionally, we demonstrate the generalization ability of MB-TAIL by extending it to the function approximation setting and proving that it can achieve expert sample and interaction complexity independent of $|\gS|$ \footnote{This paper is presented at the 39th conference on uncertainty in artificial intelligence (UAI), 2023.}.
\end{abstract}

\section{Introduction}
In real-life scenarios, sequential decision-making tasks are ubiquitous, where agents devise policies to maximize the long-term return. Reinforcement learning (RL)~\citep{sutton2018reinforcement} is a popular paradigm for learning effective policies through trial and error in unknown environments. However, RL often requires a large amount of samples and laborious reward engineering to achieve satisfactory performance in practice. Alternatively, imitation learning (IL)~\citep{argall2009survey, osa2018survey} provides a more sample-efficient approach to policy optimization by directly learning from expert demonstrations, and has been proven successful in various applications~\citep{levin16_end_to_end, shi2019taobao, jang2022bc}. By leveraging existing expert knowledge, IL methods enable efficient policy learning in situations where RL might be infeasible or expensive. Therefore, IL has become an increasingly popular and practical alternative for real-world applications.

Imitation learning (IL) is a framework that aims to minimize the difference between the expert policy and the imitated policy~\citep{ross2010efficient, xu2020error, rajaraman2020fundamental}. The two prominent IL methods are behavioral cloning (BC)~\citep{Pomerleau91bc, ross2010efficient} and adversarial imitation learning (AIL)~\citep{pieter04apprentice, syed07game,ziebart2008MEIRL, ho2016gail}. BC employs supervised learning to minimize the discrepancy between the policy distribution of the imitated policy and the expert policy. On the other hand, AIL focuses on state-action distribution matching, where the learner estimates an adversarial reward function that maximizes the policy value gap and then learns a policy to minimize the gap with the inferred reward function through a min-max optimization. Practical algorithms that build upon these principles have been developed and applied to various domains \citep{Torabi18bco, fu2018airl,  ke19imitation_learning_as_f_divergence, Kostrikov19dac, Brantley20disagreement, garg2021iqlearn, dadashi2021primal, viano2022proximal}.

A remarkable observation from empirical studies~\citep{ho2016gail, Kostrikov19dac, ghasemipour2019divergence} is that adversarial imitation learning (AIL) often outperforms behavioral cloning (BC) by a significant margin. This phenomenon has spurred numerous theoretical investigations~\citep{zhang2020gail, Chen20on_computation_and_generalization_of_gail, rajaraman2020fundamental, rajaraman2021value, xu2020error, liu2021provably, xu2022understanding} aimed at understanding the mechanisms of AIL. However, analyzing AIL is challenging because both the expert policy and environment transitions are unknown, making expert estimation and policy optimization/evaluation inaccurate. The complex min-max implementation of AIL further compounds the theoretical analysis difficulty. As a result, several prior works~\citep{pieter04apprentice, syed08lp, rajaraman2020fundamental, rajaraman2021value, xu2022understanding} have made the simplifying assumption of a known transition function to facilitate the analysis.

However, the characterization of environment transitions is often challenging in practical tasks, as noted in previous studies \citep{duan2016benchmarking, shi2019taobao}. Therefore, there has been growing interest in investigating AIL with unknown transitions, where the learner does not have prior knowledge of the transition function but can collect trajectories by interacting with the environment. This setup is widely used in empirical studies \citep{ho2016gail, fu2018airl,  ke19imitation_learning_as_f_divergence, Kostrikov19dac, Brantley20disagreement, garg2021iqlearn, li2022rethinking}. From a theoretical perspective, it is important to understand both the expert sample complexity (i.e., the number of trajectories collected by the expert) and the interaction complexity (i.e., the number of trajectories collected by the online learner) to achieve good policies, as these are of practical interest. In this paper, we investigate AIL with unknown transitions and focus on analyzing the required expert sample and interaction complexity.

Compared with the progress made in IL with known transitions, AIL with unknown transitions still lacks a well-developed theoretical foundation. Earlier works, such as FEM \citep{abbeel05exploration-and-ap} and GTAL \citep{syed07game}, estimated the transition function from expert demonstrations for imitation, rendering their algorithms impractical due to the prohibitively large expert sample complexity (as shown in \cref{table:summary-of-results}). To the best of our knowledge, the online apprenticeship learning (OAL) algorithm in \citep{shani2022online} is a promising approach that updates the policy and reward function using no-regret algorithms during environment interaction. In particular,  OAL achieves an expert sample complexity $\widetilde{\gO}(H^2|\gS|/\varepsilon^2)$  and interaction complexity $\widetilde{\gO} (H^4 |\gS|^2 |\gA| / \varepsilon^2)$\footnote{In \citep{shani2022online}, a regret $\widetilde{\gO}(\sqrt{H^4 |\gS|^2 |\gA|K} + \sqrt{H^3 |\gS| |\gA| K^2 /m})$ is proved, where $K$ is the number of interaction episodes and $m$ is the number of expert trajectories. We convert this regret guarantee to the sample complexity guarantee (see Appendix \ref{sec:from_regret_to_pac}).}, where $|\gS|$ and $|\gA|$ are the state and action space sizes, $H$ is the planning horizon, and $\varepsilon = V^{\piE} - V^{\pi}$ is the desired imitation gap. However, even with infinite environment interactions, OAL's expert sample complexity is sub-optimal, as the best expert sample complexity in the known transition setting is $\widetilde{\gO}(H^{3/2}|\gS|/\varepsilon)$ \citep{rajaraman2020fundamental}. Thus, improving AIL with unknown transitions is a significant area of research.

\begin{table}[htbp]
\centering
\caption{Expert sample complexity and interaction complexity of BC \citep{rajaraman2020fundamental}, FEM \citep{pieter04apprentice}, GTAL \citep{syed07game}, OAL \citep{shani2022online}, and MB-TAIL (ours) with unknown expert and transitions. We use $\widetilde{\gO}$ to hide logarithmic factors.}
\label{table:summary-of-results}
\begin{tabular}{@{}c|c|c@{}}
\toprule
 & \begin{tabular}[c]{@{}l@{}}Expert Sample \\ Complexity\end{tabular} & \begin{tabular}[c]{@{}l@{}}Interaction \\ Complexity\end{tabular} \\ \midrule
BC   & $\widetilde{\gO} \lp \frac{H^2 |\gS|}{\varepsilon} \rp$  & 0 \\
 FEM  & $\widetilde{\gO} \lp \frac{H^{2} |\gS|}{\varepsilon^2} + \frac{H^8 |\gS|^3 |\gA| }{\varepsilon^5} \rp $  & 0  \\
 GTAL & $\widetilde{\gO} \lp \frac{H^{2} |\gS|}{\varepsilon^2} + \frac{H^6 |\gS|^3 |\gA| }{\varepsilon^3} \rp$  & 0  \\
 OAL  & $\widetilde{\gO} \lp \frac{H^{2} |\gS|}{\varepsilon^2} \rp$  & $\widetilde{\gO} \lp \frac{H^4 |\gS|^2 |\gA| }{\varepsilon^2} \rp$ \\  \hline
 MB-TAIL  &  $\widetilde{\gO} \lp \frac{H^{3/2} |\gS|}{\varepsilon} \rp$ & $\widetilde{\gO}\lp \frac{H^3 |\gS|^2 |\gA| }{\varepsilon^2} \rp$   \\ 
 \bottomrule
\end{tabular}
\end{table}

\textbf{Contribution.} This paper presents a new and general framework (\cref{algo:framework}) that overcomes the challenge of unknown transitions and unknown expert policies. At a high level, our framework establishes a connection between AIL and reward-free exploration (RFE) \citep{chi20reward-free, menard20fast-active-learning, chen2022rewardfree}, which is an emerging topic in online RL. We prove that any effective AIL algorithm that works with known transitions can be transferred to the unknown transition setting using an efficient RFE method, as shown in \cref{prop:connection}.

Further, we also introduce a new algorithm called MB-TAIL\footnote{MB-TAIL stands for model-based transition-aware adversarial imitation learning.}, which incorporates recent advances in AIL with known transitions and RFE. MB-TAIL builds on MIMIC-MD \citep{rajaraman2020fundamental} and RF-Express \citep{menard20fast-active-learning} but requires new designs to apply their main ideas in the unknown transition setting. Notably, MB-TAIL achieves an expert sample complexity of $\widetilde{\gO}(H^{3/2}|\gS|/\varepsilon)$, meeting the lower bound $\Omega(H^{3/2}/\varepsilon)$ \citep{nived2021provably} in $H$ and $\varepsilon$. This sample complexity is nearly minimax optimal and the first to be achieved in the unknown transition setting. Additionally, MB-TAIL has an interaction complexity of $\widetilde{\gO} (H^3 |\gS|^2 |\gA|  / \varepsilon^2 )$, which improves upon the best-known OAL algorithm by a factor of $\gO(H)$.

Finally, we extend the MB-TAIL algorithm to the function approximation setting and demonstrate its ability to achieve the expert sample and interaction complexity independent of the state space size $|\gS|$. Specifically, we investigate the case of state abstraction \citep{li2006towards}, which involves approximating functions using piecewise constant functions. By employing appropriate state abstractions, MB-TAIL can estimate the abstract state-action distribution instead of the tabular counterpart, which is crucial for generalization.

\section{Related Work}
In the realm of AIL with known transitions, there have been numerous theoretical investigations into expert sample complexity \citep{pieter04apprentice, syed07game, Zahavy20al_via_frank-wolfe, rajaraman2020fundamental, swamy2022minimax, xu2021error, xu2022understanding}. For example, FEM and GTAL, which are traditional AIL algorithms, have expert sample complexity of $\widetilde{\gO} (H^2|\gS|/\varepsilon^2)$ \footnote{Results from \citep{pieter04apprentice,syed07game} are transformed from the infinite-horizon setting to the episodic setting by 1) substituting the effective planning horizon $1 / (1-\gamma)$ with the finite planning horizon $H$; 2) instantiating the linear feature with the one-hot feature under the tabular setting.}. This upper bound is proven to be tight in the worst-case \citep{xu2022understanding, swamy2022minimax}. Additionally, \citet{rajaraman2020fundamental} proposed a novel AIL technique, MIMIC-MD, which leverages the transition function to obtain an enhanced expert sample complexity of $\widetilde{\gO} (H^{3/2} |\gS| /\varepsilon)$. MIMIC-MD meets the information-theoretic lower bound of expert sample complexity with known transitions, which is $\widetilde{\Omega} (H^{3/2} /\varepsilon)$ \citep{nived2021provably}, in terms of both $H$ and $\varepsilon$. Recently, horizon-free expert sample complexity was studied in \citep{xu2022understanding}, which explains the superior performance of AIL with known transitions. However, there are only a limited number of theoretical investigations into AIL with unknown transitions. We have already discussed these in the previous section and thus will not repeat them here.

Our research establishes a connection between adversarial imitation learning and reward-free exploration, which is an emerging area of interest in online reinforcement learning. The reward-free exploration framework was introduced in \citep{chi20reward-free} with two primary goals: 1) isolating the exploration and planning problems within a standard RL framework and 2) learning an environment that is robust enough to cover all possible training scenarios. Since then, several advances have been made in this field \citep{Kaufmann21adaptive-rfe, wang2020rewardfree, zhang2021reward, chen2022rewardfree}. Specifically, \citep{menard20fast-active-learning} achieved the minimax rate in the tabular setting.

It is worth noting that AIL is closely related to inverse reinforcement learning (IRL) \citep{ng00irl}, which aims to infer the ground truth reward function from expert demonstrations. Recent works in IRL include \citep{metelli2021provably}, which studied the error propagation of the obtained policy's performance when transferring the reward function to a new environment, and \citep{zeng2022maximum}, which developed a single-loop algorithm to recover the reward function under the maximum entropy IRL formulation. Additionally, \citep{lindner2022active} proposed an upper confidence approach that actively explores the environment and expert policy to learn the reward function. However, our focus differs from these studies as our goal is to solve the imitation learning problem by learning a high-quality policy, rather than inferring the reward function.

\section{Background}

\textbf{Episodic Markov Decision Process.} In this paper, we consider episodic Markov decision process (MDP), which can be described by the tuple $\gM = (\gS, \gA, P, r, H, \rho)$. Here $\gS$ and $\gA$ are the state and action space, respectively. $H$ is the planning horizon and $\rho$ is the initial state distribution. $P = \{P_1, \cdots, P_{H}\}$ specifies the non-stationary transition function of this MDP; concretely, $P_h(s_{h+1}|s_h, a_h)$ determines the probability of transiting to state $s_{h+1}$ conditioned on state $s_h$ and action $a_h$ at time step $h$, for $h \in [H]$, where $[x]$ denotes the set of integers from $1$ to $x$. Similarly, $r = \{r_1, \cdots, r_{H}\}$ specifies the reward function of this MDP; without loss of generality, we assume that $r_h: \gS \times \gA \rar [0, 1]$, for $h \in [H]$. A non-stationary policy $\pi = \lb \pi_1, \cdots, \pi_h \rb$ with $\pi_h: \gS \rar \Delta(\gA)$, where $\Delta(\gA)$ is the probability simplex and $\pi_h (a|s)$ gives the probability of selecting action $a$ on state $s$ at time step $h$, for $h \in [H]$. 

The sequential decision process runs as follows: at the beginning of an episode, the environment is reset to an initial state according to $\rho$; then the agent observes a state $s_h$ and takes an action $a_h$ based on $\pi_h(a_h|s_h)$; consequently, the environment makes a transition to the next state $s_{h+1}$ according to $P_h(s_{h+1}|s_h, a_h)$ and sends a reward $r_h(s_h, a_h)$ to the agent. This episode ends after $H$ repeats. 

The quality of a policy is measured by its \emph{policy value} (i.e., the expected long-term return): 
\begin{align*}
    V^{\pi} = \expect \bigg[ &\sum_{h=1}^{H} r_h(s_h, a_h) | s_1\sim \rho; a_h \sim \pi_h (\cdot|s_h), s_{h+1} \sim P_h(\cdot|s_h, a_h), \forall h \in [H] \bigg].
\end{align*}
To facilitate later analysis, we introduce the state-action distribution induced by a policy $\pi$:
\begin{align*}
    d_h^{\pi}(s, a) = \sP ( s_h = s, a_h = a | s_1 \sim \rho; a_\ell \sim \pi_h (\cdot|s_\ell),
    s_{\ell+1} \sim P_{\ell} (\cdot|s_{\ell}, a_{\ell}),\; \forall \ell \in [h] ).
\end{align*}
In other words, $d_h^{\pi}(s, a)$ qualifies the visitation probability of state-action pair $(s, a)$ at time step $h$. In this way, we get an equivalent dual form of the policy value \citep{puterman2014markov}:
\begin{align}   \label{eq:dual_of_policy_value}
    V^{\pi} = \sum_{h=1}^{H} \sum_{(s, a) \in \gS \times \gA} d_h^{\pi}(s, a) r_h(s, a),
\end{align}
which will be used in later analysis.

\textbf{Imitation Learning.} The goal of IL is to learn a high quality policy \emph{without} the environment reward function. To this end, we often assume there is a nearly optimal expert policy $\piE$ that could interact with the environment to generate a dataset (i.e., $m$ trajectories of length $H$):
\begin{align*}
    \gD =& \{ \tr = \lp s_1, a_1, s_2, a_2, \cdots, s_H, a_H \rp; s_1 \sim \rho; a_h \sim \piE_h(\cdot|s_h), s_{h+1} \sim P_h(\cdot|s_h, a_h), \forall h \in [H] \}.
\end{align*}
Then, the learner can use the dataset $\gD$ to mimic the expert and to obtain a good policy. The quality of imitation is measured by the \emph{imitation gap}~\citep{pieter04apprentice, ross2010efficient, rajaraman2020fundamental}: $V^{\piE} - V^{\pi}$, where $\pi$ is the learned policy. That is, we hope the learned policy can perfectly imitate the expert such that the imitation gap is small. In this paper, we assume the expert policy is deterministic, which is common in the literature \citep{rajaraman2020fundamental, swamy2022minimax, xu2022understanding}. 

\textbf{Notation.} We denote $\Pi$ as the set of all stochastic policies for the learner. Furthermore, $|\gD|$ is the number of trajectories in $\gD$. We reserve the symbol $m$ to denote the number of expert trajectories. We write $a(n) \gtrsim b(n)$ if there exist constants $C > 0, n_0 \geq 1$ such that  $a(n) \geq Cb(n)$ for $n \geq n_0$.

\section{Warm-up: AIL with Known Transitions}
\label{sec:warm-up}

To imitate the expert policy, AIL methods solve the state-action distribution matching problem \citep{ho2016gail, ke19imitation_learning_as_f_divergence, xu2020error}. As an introduction to general readers, we consider the known transition setting in this section. Our starting point is the following state-action distribution matching problem: 
\begin{align} \label{eq:ail_known_transition} 
   \min_{\pi \in \Pi}  \sum_{h=1}^{H} \lnorm d^{\pi}_h - \widetilde{d}^{\piE}_h \rnorm_1.
\end{align}
where $\widetilde{d}^{\piE}_h$ is an estimation of the expert state-action distribution $d^{\piE}_h$.  We can explain why \cref{eq:ail_known_transition} is a good learning objective with the following two definitions.

\begin{defn}
\label{def:estimation}
An estimator $\widetilde{d}^{\piE}_h$ is said to be $\varepsilon_{\est}$-accurate for $d^{\piE}_h$ if 
\begin{align*}
    \sum_{h=1}^{H} \lnorm \widetilde{d}^{\piE}_h - d^{\piE}_h \rnorm_1 \leq \varepsilon_{\est}.  
\end{align*}
\end{defn}

\begin{defn}
\label{def:distribution_matching_error}
For optimization problem \eqref{eq:ail_known_transition}, a policy $\widebar{\pi}$ is said to be $\varepsilon_{\opt}$-optimal if 
\begin{align*}
    \sum_{h=1}^{H} \lnorm d^{\widebar{\pi}}_h - \widetilde{d}^{\piE}_h \rnorm_1 \leq \min_{\pi \in \Pi}  \sum_{h=1}^{H} \lnorm d^{\pi}_h - \widetilde{d}^{\piE}_h \rnorm_1 + \varepsilon_{\opt}. 
\end{align*}
\end{defn}

\begin{lem}  \label{lem:1}
Given an $\varepsilon_{\est}$-accurate estimator $\widetilde{d}^{\piE}_h$, suppose that $\widebar{\pi}$ is $\varepsilon_{\opt}$-optimal for problem \eqref{eq:ail_known_transition}, then we have that $V^{\piE} - V^{\widebar{\pi}} \leq  \varepsilon_{\opt} + 2\varepsilon_{\est}$.
\end{lem}

Proof of \cref{lem:1} can be found in the Appendix along with other theoretical results. This lemma establishes a strong theoretical foundation for state-action distribution matching. It is worth noting that similar versions of this lemma have been presented in prior works such as \citep{syed07game, rajaraman2020fundamental}. We will discuss how to control estimation and optimization errors in the next section.

While significant theoretical progress has been made in the known transition setting, this assumption is not always practical in real-world applications where the transition function is unknown. In such cases, empirical studies have been carried out under the unknown transition setting, where the interaction with environments is allowed but the analytic form of transition function is not available. In addition to expert sample complexity, the interaction complexity is also of great interest in this scenario, which we will explore in the next section.

\section{Main Results: AIL with Unknown Transitions}
\label{sec:main_result}

In this section, we consider the unknown transition setting where $d^{\pi}_h$ is not accessible, rendering the learning objective in \cref{eq:ail_known_transition} inapplicable. A sound solution is to replace $d^{\pi}_h$ with its estimated version $\widehat{d}^{\pi}_h$ in \cref{eq:ail_known_transition}. We highlight that the unknown transition leads to the exploration-and-exploitation trade-off, which is shared with online RL \citep{agarwal2022rlbook}. The prior work OAL addresses this challenge by an optimistic estimation of the value function \citep{shani2022online}.

In this paper, we explore an alternative model-based approach: we first learn the transition function from collected trajectories and subsequently estimate $d^{\pi}_h$ based on the recovered transition model. The key challenge is how to recover a good transition model such that policy evaluation/optimization can be conducted accurately. To this end, we propose a general algorithmic framework, which connects AIL with reward-free exploration (or RFE for short)~\citep{chi20reward-free, menard20fast-active-learning}, which is an emerging topic in online RL. Under this framework, a proper AIL algorithm that works under the known transition setting could be transferred to the unknown transition setting by leveraging an efficient RFE method. Before presenting the details of our framework, we formally introduce RFE.

\begin{defn}[\citep{menard20fast-active-learning}]   \label{defn:reward_free}
Given an MDP $\gM$ without reward function $r$, an algorithm is said to be $(\varepsilon, \delta)$-PAC for reward-free exploration (RFE) if 
\begin{align*}
    \sP \big( &\text{for any reward function $r$}, |V^{\pi^*_{r}} - V^{\widehat{\pi}_{r}^*}| \leq \varepsilon \big) \geq 1 - \delta,
\end{align*}
where $\pi^*_{r}$ is the optimal policy in the MDP with the reward function $r$, and $\widehat{\pi}_{r}^*$ is the optimal policy in the MDP with the learned transition model $\widehat{P}$ by RFE and reward function $r$.
\end{defn}

By algorithmic designs, RFE methods usually satisfy the so-called uniform policy evaluation property, which is crucial for the discussion of AIL.
\begin{defn}  \label{defn:uniform_policy_evaluation}
Given an MDP $\gM$ without reward function $r$, an algorithm is said to be $(\varepsilon, \delta)$-PAC for uniform policy evaluation if 
\begin{align*}
    \sP \big( \text{for any reward function $r$ and policy $\pi$}, |V^{\pi, P, r} - V^{\pi, \widehat{P}, r}| \leq \varepsilon \big) \geq 1 - \delta,
\end{align*}  
where $V^{\pi, P, r}$ and $V^{\pi, \widehat{P}, r}$ are the policy values of policy $\pi$ with reward function $r$ under the real transition model $P$ and recovered transition model $\widehat{P}$, respectively.
\end{defn}

Examples of algorithms that satisfy \cref{defn:uniform_policy_evaluation} include RF-RL-Explore \citep{chi20reward-free} (see their Lemma 3.6), RF-UCRL \citep{Kaufmann21adaptive-rfe} (see their Lemma 1 and the stopping rule) and RF-Express in \citep{menard20fast-active-learning} (see their Lemma 1 and the stopping rule).

\cref{defn:uniform_policy_evaluation} is connected with AIL in the following way:
\begin{align*}
 \sum_{h=1}^H \| \widehat{d}^{\pi}_h -  d^{\pi}_h  \|_1 =\max_{w \in \gW} \sum_{h=1}^H \sum_{(s, a)} w_h (s, a) (\widehat{d}^{\pi}_h (s, a) -  d^{\pi}_h (s, a) ) = \max_{w \in \gW} V^{\pi, \widehat{P}, w} - V^{\pi, P, w} \leq \varepsilon.
\end{align*}
Here the first equality follows the dual representation of $\ell_1$-norm, and $\gW = \{  w: \Vert w \Vert_{\infty} \leq 1\}$ is the unit ball. The second equality follows \cref{eq:dual_of_policy_value}. The last inequality follows \cref{defn:uniform_policy_evaluation}. In plain language, the above formula shows that we can get an accurate estimation of $d^{\pi}_h$, based on the recovered model by RFE.

Based on the above relation, with a transition model learned by RFE, AIL can be implemented as if this empirical transition function were the same as the true transition function. More specifically, the state-action distribution matching problem \cref{eq:ail_known_transition} becomes
\begin{align}   \label{eq:ail_with_model}
 \min_{\pi \in \Pi} \sum_{h=1}^{H} \lnorm  \widetilde{d}^{\piE}_h -  d^{\pi, \widehat{P}}_h \rnorm_1 
\end{align}
where $d^{\pi, \widehat{P}}_h$ is the state-action distribution of policy $\pi$ with the transition model $\widehat{P}$. We outline the whole procedure in \cref{algo:framework} and the theoretical guarantee is provided below.

\begin{prop}   \label{prop:connection}
Suppose that 
\begin{itemize}  %
    \item[(a)] a reward-free exploration algorithm A satisfies the uniform policy evaluation property (see \cref{defn:uniform_policy_evaluation}) up to an error $\varepsilon_{\rfe}$ with probability at least $1-\delta_{\rfe}$;
    \item[(b)] an algorithm B has a state-action distribution estimator for $d^{\piE}_h$, which satisfies $\sum_{h=1}^H \Vert \widetilde{d}^{\piE}_h - d^{\piE}_h  \Vert_{1} \leq \varepsilon_{\est}$, with probability at least $1-\delta_{\est}$;
    \item[(c)] with the transition model in (a) and the estimator in (b), an algorithm C solves the optimization problem in \cref{eq:ail_with_model} up to an error $\varepsilon_{\opt}$.
\end{itemize}
Then applying algorithms A, B and C under the framework in Algorithm \ref{algo:framework} could return a policy $\widebar{\pi}$, which has a policy value gap (i.e., $V^{\piE} - V^{\widebar{\pi}}$) at most $2 \varepsilon_{\est} + 2 \varepsilon_{\rfe} + \varepsilon_{\opt}$, with probability at least $1-\delta_{\est} - \delta_{\rfe}$.
\end{prop}

\begin{algorithm}[htbp]
\caption{Meta-algorithm for AIL with Unknown Transitions}
\label{algo:framework}
\begin{algorithmic}[1]
\REQUIRE{Expert demonstrations $\gD$.}
\STATE{$\widehat{P} \lar$ Invoke a reward-free exploration method to collect $n$ trajectories and learn a transition model.}
\STATE{$\widetilde{d}_h^{\piE} \lar $ Estimate the expert state-action distribution.}
\STATE{$\widebar{\pi} \lar$ Apply an AIL approach to perform imitation with the expert estimation $\widetilde{d}_h^{\piE}$ under transition model $\widehat{P}$.}
\ENSURE{Policy $\widebar{\pi}$.}
\end{algorithmic}
\end{algorithm}

Next, we show how to substantiate the framework in \cref{algo:framework} with detailed procedures. We will consider the tabular formulation, where the space of parameterized value functions spans all possible functions. In this scenario, expert policies and reward functions are realizable. We discuss how to control $\varepsilon_{\rfe}$, $\varepsilon_{\est}$, and $\varepsilon_{\opt}$ in a sequential order.

\subsection{Controlling Reward-free Exploration Error}

To ensure that condition (a) in \cref{prop:connection} is satisfied, we make use of the RF-Express algorithm, as described in \citep{menard20fast-active-learning}. This advanced algorithm allows us to control $\varepsilon_{\rfe}$ effectively. Below, we provide the theoretical property of RF-Express.

\begin{lem}[Theorem 1 in \citep{menard20fast-active-learning}]
\label{lem:reward_free}
Fix $\varepsilon \in \lp 0, 1 \rp$ and $\delta \in (0, 1)$. Consider the \textnormal{RF-Express} algorithm (see Algorithm \ref{algo:rf_express} in Appendix) and $\widehat{P}$ is the empirical transition function built on the collected trajectories, if the number of trajectories collected by \textnormal{RF-Express} satisfies 
\begin{align*}
    n \gtrsim  \frac{H^{3} |\gS| |\gA| }{\varepsilon^2}    \lp |\gS| + \log\lp\frac{|\gS| H}{\delta} \rp \rp.
\end{align*}
Then with probability at least $1-\delta$, for any policy $\pi$ and any bounded reward function $r$ between $[-1, 1]$, we have $| V^{\pi, P, r} - V^{\pi, \widehat{P}, r} | \leq {\varepsilon}/{2}$; furthermore, for any bounded reward function $r$ between $[-1, 1]$, we have $ \max_{\pi \in \Pi} V^{\pi, P, r} \leq V^{\widehat{\pi}_{r}^{*}, P, r} + \varepsilon$, where $\widehat{\pi}_{r}^{*}$ is the optimal policy under the empirical transition function $\widehat{P}$ with reward function $r$.

\end{lem}

\subsection{Controlling Expert State-action Distribution Estimation Error}
\label{subsec:transition_aware_estimator}

In this part, we talk about how to control the expert state-action distribution estimation error. Quite often, the maximum likelihood estimator (MLE) is considered in the literature \citep{pieter04apprentice, syed07game, shani2022online}. Mathematically, MLE counts how frequently a state-action pair appears in the observed expert trajectories: 
\begin{align}   \label{eq:estimate_by_count}
    \widehat{d}^{\piE}_h(s, a) = \frac{  \sum_{\tr \in \gD}  \indict\lb \tr_h(\cdot, \cdot) = (s, a) \rb }{|\gD|}, 
\end{align}
where $\tr_h(\cdot, \cdot)$ indicates the specific state-action pair of trajectory $\tr$ in time step $h$. The sample complexity of MLE is well-known. 
\begin{lem}[\cite{rajaraman2020fundamental}]   \label{lem:sample_complexity_mle}
Fix $\varepsilon \in (0, H)$ and $\delta \in (0, 1)$, if the number of expert trajectories in $\gD$ satisfies
\begin{align*}
    m \gtrsim \frac{H^2 \vert \gS \vert}{\varepsilon^2} \log \lp \frac{H}{\delta} \rp,
\end{align*}
then with probability at least $1-\delta$, we have $\sum_{h=1}^H \Vert \widehat{d}^{\piE}_h - d^{\piE}_h   \Vert_{1} \leq \varepsilon$.
\end{lem}

{The above sample complexity of MLE is tight in the worst case; see, e.g., \citep[Lemma 8]{kamath2015learning}. Though MLE can be implemented under our framework, this estimator cannot lead to the minimax optimal expert sample complexity $\Theta(H^{3/2}|\gS|/\varepsilon)$.} To address this issue, in light of \citep{rajaraman2020fundamental}, we develop a new estimator. For a better presentation, let us introduce the following notations.
\begin{itemize}  %
    \item Similar to $\tr_h(\cdot, \cdot)$,  $\tr_h(\cdot)$ indicates the specific state of trajectory $\tr$ in time step $h$.
    \item Without $(\cdot)$ or $(\cdot, \cdot)$, $\tr_h$ is the truncated version of trajectory $\tr$ up to time step $h$, i.e., $\tr_h = (s_1, a_1, \cdots, s_h, a_h)$.
    \item $\gS_{h}(\gD) = \{s: \exists \tr \in \gD \text{ such that } s = \tr_h(\cdot) \}$ is the set of states visited at time step $h$ in $\gD$. 
    \item $\Tr_h^{\gD} = \{ \tr_h = (s_1, a_1, \ldots, s_h, a_h): s_\ell \in \gS_{\ell}(\gD), \forall \ell \in [h] \}$ is the set of truncated  trajectories (that may not appear in $\gD$), along which each state has been visited in $\mathcal{D}$ up to time step $h$.
\end{itemize}

From the definition of state-action distribution, we have  
\begin{align}
   d^{\pi}_h(s, a)  = d^{\pi}_h(s) \pi_h(a|s) =  \big[ \sum_{s^\prime, a^\prime} d^{\pi}_{h-1}(s^\prime, a^\prime) P_{h-1}(s|s^\prime, a^\prime) \big]  \pi_h(a|s)  \label{eq:flow}
\end{align}
This equation offers another perspective on visitation probability: $d^{\pi}_h(s, a)$ represents the weighted average of flows. Specifically, each flow path is determined by ancestral state-action sequences that lead to the target state-action pair $(s, a)$, and the weight of this flow is influenced by both the transition probability and the policy distribution.

However, when dealing with a finite sample regime, only a subset of trajectories executed by the expert policy is observed, while others remain unobserved. We can use the transition function to calculate the visitation probability for the observed trajectories, but we require statistical estimation for the non-observed ones. This idea has been exploited in \citep{rajaraman2020fundamental} in the known transition setting.

Now, consider the dataset $\gD$ is randomly divided into two equal parts, i.e., $\gD = \gD_1 \cup \gD_1^{c}$ and $\gD_1 \cap \gD_1^{c} = \emptyset$ with $|\gD_1 |=  |\gD_1^{c}| = m / 2$. We have the following decomposition:
\begin{align} 
d_h^{\piE}(s, a) =  \underbrace{\sum_{\tr_h \in \Tr_h^{\gD_1}} \sP^{\piE}(\tr_h) \indict\lb \tr_h(\cdot, \cdot) = (s, a) \rb}_{:= \clubsuit} + \underbrace{\sum_{\tr_h \notin \Tr_h^{\gD_1}} \sP^{\piE}(\tr_h) \indict\lb \tr_h(\cdot, \cdot) = (s, a) \rb}_{:= \spadesuit}, \label{eq:key_decomposition}
\end{align}
where $\sP^{\piE} (\tr_h)$ is the probability of the truncated trajectory $\tr_h$ induced by the deterministic expert policy $\piE$. As we have mentioned, if the transition function is known, we can calculate  $\sP^{\piE} (\tr_h)$ directly: $\sP^{\piE} (\tr_h)  = \rho(s_1) \prod_{\ell=1}^{h-1} P_{\ell}(s_{\ell + 1}|s_{\ell}, a_{\ell})$ with $\tr_h = (s_1, a_1, \cdots, s_h, a_h)$

We explain two terms in \cref{eq:key_decomposition} separately. On the one hand, term $\clubsuit$ can be calculated exactly if we know both the transition function and $\gD_1$, as explained previously. However, this is not applicable in our case as the transition function is unknown. We will discuss how to deal with this trouble later. On the other hand, term $\spadesuit$ accounts for non-observed trajectories, which is not easy to compute (because we have no clue about expert actions on non-observed states). To address this issue, \citet{rajaraman2020fundamental} proposed to use trajectories in $\gD_1^{c}$ to make a maximum likelihood estimation. This is because, $\gD_1^{c}$ is statistically independent of $\gD_1$ and therefore can be viewed as a new dataset. We follow the approach in \citep{rajaraman2020fundamental} to estimate term $\spadesuit$.

Now, we explain how to estimate term $\clubsuit$ in the unknown transition setting. Our solution has two steps. The first step is to apply BC on $\gD_1$ to learn policy $\pi^\prime$:
\begin{align*}
\pi^{\prime}_h(a|s) = \left\{ \begin{array}{cc}
  \frac{n^1_h(s, a)}{n^1_h (s)}   &  \text{ if } {n^1_h(s) > 0} \\
   \frac{1}{|\gA|}  & \text{ otherwise}  
\end{array} \right.
\end{align*}
Here $n^1_h(s, a)$ ($n^1_h(s)$) is the number of state-action (state) pairs that appeared in $\gD_1$ in step $h$. This step recovers the expert behaviors on visited states in $\gD_1$. The second step is to let $\pi^\prime$ interact with the environment to collect a new dataset $\gD_{\env}^\prime$, from which we can estimate term $\clubsuit$ by MLE. To get a better sense, we mention that the uncertainty of estimating term $\clubsuit$ comes from the transition function, rather than the expert policy. Furthermore, by our design, trajectories in $\gD^{\prime}_{\env}$ are collected as if the expert policy were roll-out (because $\pi^\prime$ can perfectly match $\piE$ on $\gS(\gD_1)$, see \cref{lemma:unknown-transition-unbiased-estimation} in Appendix for more details), so the randomness of MLE is only caused by the stochastic transitions.

In summary, we arrive at the following estimator:
\begin{align}
    \widetilde{d}^{\piE}_h(s, a) = {\frac{\sum_{\tr_h \in \gD_{\env}^\prime} \indict \{ \tr_h (\cdot, \cdot) = (s, a), \tr_h \in \Tr_h^{\gD_1} \}}{|\gD^\prime_{\env}|}} + {\frac{  \sum_{\tr_h \in \gD_1^c}  \indict\{ \tr_h (\cdot, \cdot) = (s, a), \tr_h \not\in \Tr_h^{\gD_1}  \} }{|\gD_1^c|}}. \label{eq:new_estimator_unknown_transition}
\end{align}
Two terms in \cref{eq:new_estimator_unknown_transition} give estimation for terms $\clubsuit$ and $\spadesuit$ in \cref{eq:key_decomposition}, respectively. {It is important to note that the state-action distribution largely depends on the transition probability, as shown in \cref{eq:flow}. In contrast to the MLE in \cref{eq:estimate_by_count}, our proposed estimator additionally leverages the transition information from the online interactions; see the first term in RHS in \cref{eq:new_estimator_unknown_transition}. This advancement leads to a more accurate estimation of the expert's state-action distribution.} 
\begin{lem} \label{lemma:sample_complexity_of_new_estimator_unknown_transition}
Given the expert dataset $\gD$, let $\gD$ be divided into two equal subsets, i.e., $\gD = \gD_{1} \cup \gD_{1}^c$ and $\gD_1 \cap \gD_1^{c} = \emptyset$ with $\labs \gD_1 \rabs = \labs \gD_1^{c} \rabs = m / 2$. Fix $\pi^{\prime} \in \Pi_{\text{BC}} \lp \gD_1 \rp$, let $\gD^\prime_{\mathrm{env}}$ be the dataset collected by $\pi^\prime$ and $|\gD^\prime_{\mathrm{env}} | = n^\prime$. Fix $\varepsilon \in (0, 1)$ and $\delta \in (0, 1)$; suppose $H \geq 5$. Consider the estimator $\widetilde{d}^{\piE}_h$ shown in \eqref{eq:new_estimator_unknown_transition}, if the expert sample complexity ($m$) and the interaction complexity ($n^\prime$) satisfy
\begin{align*}
    m \gtrsim   \frac{H^{3/2} | \gS | }{\varepsilon} \log\lp  \frac{|\gS| H}{\delta} \rp, \; n^\prime \gtrsim \frac{H^{2} | \gS |}{\varepsilon^2} \log\lp  \frac{|\gS| H}{\delta} \rp,
\end{align*}
then with probability at least $1-\delta$, we have
\begin{align*}
    \sum_{h=1}^H \lnorm \widetilde{d}^{\piE}_h - d^{\piE}_h  \rnorm_{1} \leq \varepsilon.
\end{align*}
\end{lem}

To our best knowledge,  the estimator \eqref{eq:new_estimator_unknown_transition} is the first to enjoy a better expert sample complexity than MLE in the unknown transition setting. The nature of unknown transitions raises a technical difficulty in analyzing the estimation error of two sub-estimators in \eqref{eq:new_estimator_unknown_transition}. We highlight that the classical concentration inequality, used to analyze the MLE estimator in \cref{lem:sample_complexity_mle}, cannot be used to upper bound this estimation error, as the distributions involved are not valid. To overcome this obstacle, we employ Chernoff's bound and additional statistical arguments.

\subsection{Controlling Optimization Error}

 We now consider the optimization issue. Again, we utilize the dual representation of $\ell_1$-norm and the min-max theorem~\citep{bertsekas2016nonlinear} to obtain the following max-min optimization problem:
\begin{align}   \label{eq:new_algo_max_min}
     \max_{w \in \gW} \min_{\pi \in \Pi} \sum_{h=1}^H \sum_{(s, a)} w_h (s, a) ( \widetilde{d}^{\piE}_h (s, a) -  d^{\pi, \widehat{P}}_h (s, a) ).
\end{align}
where $\gW = \{w: \lnorm w \rnorm_\infty \leq 1 \}$ is the unit ball. We see that the inner problem in \eqref{eq:new_algo_max_min} is to maximize the policy value of $\pi$ given the reward function $w_h(s, a)$ (see \cref{eq:dual_of_policy_value} for the dual form of policy value). For the outer optimization problem, we can use online gradient descent methods \citep{shalev12online-learning} so that the overall objective can finally reach an approximate saddle point. Formally, let us define the objective $f^{(t)}(w)$:
\begin{align}   
    & \underbrace{\sum_{h=1}^{H} \sum_{(s, a) \in \gS \times \gA} w_h(s, a) \lp d^{\pi^{(t)}, \widehat{P}}_h (s, a) - \widetilde{d}^{\piE}_h (s, a)  \rp}_{:= f^{(t)}(w)}, \label{eq:objective_w}
\end{align}
where $\pi^{(t)}$ is the optimized policy in iteration $t$. Then the update rule for $w$ is:
\begin{align*}   
    w^{(t+1)} := \gP_{\gW} ( w^{(t)} - \eta^{(t)}  \nabla f^{(t)}(w^{(t)}) ), 
\end{align*}
where $\eta^{(t)} > 0$ is the stepsize to be chosen later, and $\gP_{\gW}$ is the Euclidean projection on the unit  ball $\gW$, i.e., $\gP_{\gW}(w) := \argmin_{z \in \gW} \lnorm z- w \rnorm_2$. The procedure for solving \eqref{eq:new_algo_max_min} is outlined in Algorithm~\ref{algo:gradient_based_optimization}.

\begin{algorithm}[htbp]
\begin{algorithmic}[1]
\caption{Gradient-based Optimization}
\label{algo:gradient_based_optimization}
\REQUIRE{Transition model $\widehat{P}$, and expert state-action distribution estimator $\widetilde{d}^{\piE}_h$.}
\FOR{$t = 1, 2, \cdots, T$}
\STATE{$\pi^{(t)} \lar $ Solve the optimal policy with the transition model $\widehat{P}$ and reward function $w^{(t)}$ up to an error $\varepsilon_{\rl}$.}
\STATE{Compute the state-action distribution $d^{\pi^{(t)}, \widehat{P}}_h$ for $\pi^{(t)}$.}
\STATE{Update $ w^{(t+1)} := \gP_{\gW}\lp w^{(t)} - \eta^{(t)}  \nabla f^{(t)}(w^{(t)}) \rp$ with $f^{(t)}(w)$ defined in \cref{eq:objective_w}.}
\ENDFOR
\STATE{Compute the mean state-action distribution $\widebar{d}_h(s, a) = \sum_{t=1}^{T} d^{\pi^{(t)}, \widehat{P}}_h(s, a) / T$.}
\STATE{Derive $\widebar{\pi}_h (a|s) \lar \widebar{d}_h(s, a) / \sum_{a} \widebar{d}_h(s, a)$.}
\ENSURE{Policy $\widebar{\pi}$.}
\end{algorithmic}
\end{algorithm}

Line 2 in \cref{algo:gradient_based_optimization} formulates a typical reinforcement learning (RL) optimization problem. We allow $\pi^{(t)}$ to be $\varepsilon_{\rl}$-optimal with respect to the optimal policy with reward function $w^{(t)}$, i.e., $V^{\pi^{(t)}, \widehat{P}, w^{(t)}} \geq V^{\pi^{*}_{w^{(t)}}, \widehat{P}, w^{(t)}} - \varepsilon_{\rl}$. In the tabular case, $\varepsilon_{\rl} = 0$ by value iteration with finite and polynomial computation steps. For approximate methods such as policy gradient ascent, we require that they can guarantee $\varepsilon_{\rl}$ is small with low computational cost.

\begin{lem} \label{lemma:approximate-minimax}
Fix $\varepsilon > 0$. Consider the gradient-based optimization procedure in \cref{algo:gradient_based_optimization} with $\varepsilon_{\rl} \leq \varepsilon/2$.  If we take $T \gtrsim H^2 |\gS||\gA|/\varepsilon^2$ and $\eta^{(t)} := \sqrt{|\gS||\gA| / (8T)}$, then we have
\begin{align*}
\sum_{h=1}^H \lnorm d^{\widebar{\pi}, \widehat{P}}_h - \widetilde{d}^{\piE}_h \rnorm_{1} \leq \min_{\pi \in \Pi} \sum_{h=1}^H \lnorm d^{\pi, \widehat{P}}_h - \widetilde{d}^{\piE}_h \rnorm_{1} + \varepsilon. 
\end{align*}
\end{lem}

\subsection{MB-TAIL: Combing All Together}

Combing the above all pieces together, we obtain the final approach called MB-TAIL presented in Algorithm \ref{algo:mbtail-abstract}. Here \dquote{MB-TAIL} stands for model-based transition-aware adversarial imitation learning.

\begin{algorithm}[htbp]
\caption{Model-based Transition-aware AIL}
\label{algo:mbtail-abstract}
\begin{algorithmic}[1]
\REQUIRE{Expert demonstrations $\gD$.}
\STATE{Invoke \textnormal{RF-Express} to collect $n$ trajectories and learn an empirical transition function $\widehat{P}$.}
\STATE{Randomly split $\gD$ into two equal parts: $\gD = \gD_1 \cup \gD_1^{c}$.}
\STATE{Learn $\pi^{\prime} \in \Pi_{\text{BC}} \lp \gD_{1} \rp$ by BC and roll out $\pi^{\prime}$ to obtain dataset $\gD_{\env}^\prime$ with $|\gD_{\env}^\prime| = n^{\prime}$.}
\STATE{Obtain the estimator $\widetilde{d}_h^{\piE}$ in \eqref{eq:new_estimator_unknown_transition} with $\gD$ and $\gD_{\env}^\prime$.}
\STATE{$\widebar{\pi} \lar$ Apply \cref{algo:gradient_based_optimization} with the estimation $\widetilde{d}_h^{\piE}$ under transition model $\widehat{P}$.}
\ENSURE{Policy $\widebar{\pi}$.}
\end{algorithmic}
\end{algorithm}

\begin{thm}\label{theorem:sample-complexity-unknown-transition}
Fix $\varepsilon \in \lp 0, 1 \rp$ and $\delta \in (0, 1)$; suppose $H \geq 5$. Under the unknown transition setting, consider \mbalgname displayed in Algorithm \ref{algo:mbtail-abstract} and $\widebar{\pi}$ is output policy, assume that the RL error $\varepsilon_{\rl} \leq \varepsilon / 2$, the number of iterations and the step size are the same as in \cref{lemma:approximate-minimax}, if the expert sample complexity and the interaction complexity satisfy
\begin{align*}
m \gtrsim  \frac{ H^{3/2} |\gS|}{\varepsilon} \log\lp\frac{H |\gS|}{\delta} \rp, n^{\prime} \gtrsim  \frac{ H^2 |\gS|}{\varepsilon^2} \log \lp \frac{H |\gS| }{\delta} \rp, n \gtrsim \frac{H^3 |\gS| |\gA|}{\varepsilon^2} \lp |\gS| + \log \lp \frac{H |\gS| |\gA|}{\delta \varepsilon} \rp \rp,
\end{align*}
then with probability at least $1-\delta$, we have $V^{\piE} - V^{\widebar{\pi}} \leq \varepsilon $.
\end{thm}

\begin{rem}

Our MB-TAIL algorithm achieves expert sample complexity $m = \widetilde{\gO}(H^{3/2} |\gS| / \varepsilon)$ and total interaction complexity $n + n^\prime = \widetilde{\gO}(H^{3} |\gS|^2 |\gA| / \varepsilon^2)$, even in the case of unknown transitions. In comparison, the OAL algorithm in \citep{shani2022online} has expert sample complexity $\widetilde{\gO} (H^{2} |\gS| / \varepsilon^2 )$ and interaction complexity $\widetilde{\gO} ( H^4 |\gS|^2 |\gA| / \varepsilon^2 )$ in the same scenario. Theorem \ref{theorem:sample-complexity-unknown-transition} validates that our approach provides significant improvements over OAL in terms of both expert sample complexity and interaction complexity. 

The success of this improvement hinges on the design of our algorithm. Unlike OAL, which uses a maximum likelihood estimate of the expert's state-action distribution for imitation, MB-TAIL leverages transition information to construct a more accurate estimator. In addition, OAL uses a tailored optimistic value function in a model-free manner for exploration, but MB-TAIL employs a model-based, reward-free exploration method to efficiently explore the state-action space. These algorithmic designs yield substantial enhancements in both expert sample complexity and interaction complexity.

\end{rem}

\begin{figure}[htbp]
    \centering
    \begin{subfigure}{0.4\textwidth}
    \centering
    \includegraphics[width=\linewidth]{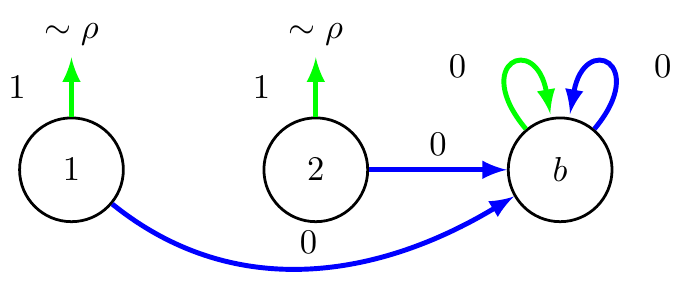}
    \caption{Reset Cliff MDP}
    \label{fig:reset_cliff}
    \end{subfigure}
    \begin{subfigure}{0.4\textwidth}
    \centering
        \includegraphics[width=\linewidth]{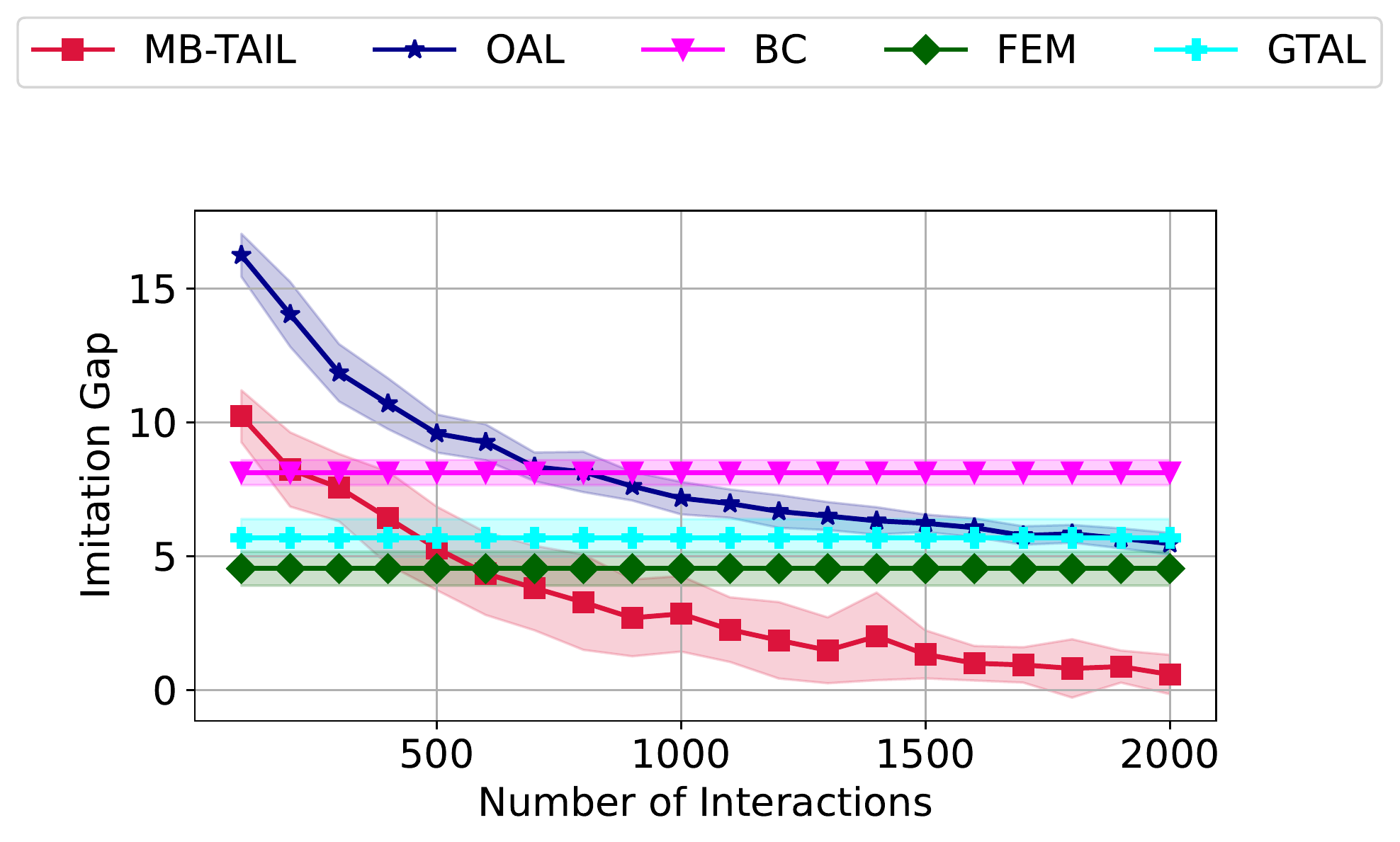}
        \caption{The imitation gap (i.e., $V^{\piE} - V^{\pi}$).}
    \end{subfigure}
    \caption{The experiment results in Reset Cliff}
    \label{fig:unknown_transition_result}
\end{figure}

\textbf{Simulation Studies.} Finally, we conclude by validating our theoretical results through experiments, where we compare the performance of MB-TAIL with four other state-of-the-art algorithms: BC \citep{Pomerleau91bc}, FEM \citep{abbeel05exploration-and-ap}, GTAL \citep{syed07game}, and OAL \citep{shani2022online}. All algorithms are given 100 expert trajectories, and we evaluate their performance on the Reset Cliff MDP, which is known to be challenging for imitation learning algorithms \citep{rajaraman2020fundamental, xu2021error}. In the Reset Cliff MDP, the state space $\gS = \{1, 2, \cdots, |\gS| - 1, b \}$ and action space $|\gA| = \{1, 2, \ldots, |\gA| - 1,  a^{\expert} \}$, where $b$ is a unique absorbing state and $a^{\expert}$ is the expert action. An example with three states and two actions is shown in Figure \ref{fig:reset_cliff}, where the expert action is shown in green. Only the expert action has a reward $+1$. All non-expert actions have the same transitions and rewards. The initial state distribution $\rho = (1/m, 1/m,  1-|\gS|/m+2/m, 0)$.

We conduct experiments with $20$ random seeds, and provide more experimental details in Appendix \ref{section:experiment_details}. The code to reproduce our results is available at our GitHub repository \footnote{\url{https://github.com/tianxusky/tabular-ail}}.

Figure \ref{fig:unknown_transition_result} shows the imitation gap for each algorithm. Note that BC, FEM, and GTAL do not leverage environment interactions. Our results show that MB-TAIL outperforms FEM and GTAL when the number of interactions exceeds $500$. Additionally, we observe that MB-TAIL outperforms OAL with the same number of interactions, which confirms the superior theoretical bounds of MB-TAIL.

\section{MB-TAIL with State Abstraction}
\label{sec:mbtail_state_abstraction}

Previously, we considered the tabular representation, which leads to theoretical bounds that depend on the size of the problem $|\gS|$. However, as suggested by the lower bounds in \citep[Theorem 6.1, 6.2]{rajaraman2020fundamental}, this dependence is inevitable and could be unacceptable when $|\gS|$ is huge. In this section, we investigate the use of state abstractions \citep{li2006towards} within MB-TAIL, so the dependence on $|\gS|$ can be eliminated.

\subsection{AIL Meets State Abstraction}
We assume that we have a set of state abstractions $\{ \phi_h \}_{h=1}^H$, where $\phi_h: \gS \rightarrow \Phi$ and $\Phi$ is the abstract state space. State abstractions correspond to function approximations using a series of piecewise constant functions \citep{chen2019information}. The abstract state space is much smaller than the original state space, i.e., $|\Phi| \ll |\gS|$. We also assume that $\{ \phi_h \}_{h=1}^H$ satisfies a condition that is common in the literature \citep{li2006towards, jiang2015abstraction}.

\begin{asmp}
\label{asmp:state_abstraction}
There exists a set of known state abstractions $\{ \phi_h \}_{h=1}^H$, which satisfies $\forall h \in [H]$, for any $s^{1}, s^{2} \in \gS$ such that $\phi_h (s^{1}) = \phi_h (s^{2})$,
\begin{align}
    &\text{bisimulation}: \forall a \in \gA, x^\prime \in \Phi, r_h (s^{1}, a) = r_h (s^{2}, a), \sum_{s^\prime \in \phi_h^{-1} (x^\prime)} P_h (s^\prime |s^1, a) = \sum_{s^\prime \in \phi_h^{-1} (x^\prime)} P_h (s^\prime |s^2, a); \label{eq:bisimulation}
    \\
    &\piE\text{-irrelevant}: \; \piE_h (s^1) = \piE_h (s^2), \label{eq:expert_consistent}
\end{align}
where $\phi_h^{-1} (x^\prime) = \{ s^\prime \in \gS: \phi_h (s) = x^\prime \}$.
\end{asmp}

{In bisimulation, the reward-consistent condition in \eqref{eq:bisimulation} ensures that two different states mapped to the same abstract state share an identical reward. We highlight that this condition is important for MB-TAIL to avoid the dependence of \emph{expert sample complexity} on $\vert \gS \vert$.} In particular, the bottleneck of the sample complexity of AIL methods is the estimation of $d^{\piE}_h(s, a)$. Under the reward-consistent condition, we can calculate the expert policy value as
\begin{align*}
    V^{\piE} = \sum_{h=1}^H \sum_{(s, a) \in \gS \times \gA} r_h (s, a)  d^{\piE}_h (s, a) = \sum_{h=1}^H \sum_{(x, a) \in \Phi \times \gA} r^{\phi}_h (x, a) d^{\piE, \phi}_h (x, a), 
\end{align*}
where $r^{\phi}_h (x, a) = r_h (s, a)$ for an arbitrary $s \in \phi^{-1}_h (x)$ and $d^{\pi, \phi}_h (x, a) = \sP^{\piE} (\phi_h (s_h) = x, a_h = a) = \sum_{s \in \phi_h^{-1} (x)} d^{\pi}_h (s, a)$ is the {abstract state-action distribution}. With the above formulation, to estimate the expert policy value, we can estimate the \emph{abstract} state-action distribution rather than the tabular counterpart, which can remove the dependence on $\vert \gS \vert$. {Analogously, the transition-consistent condition in \eqref{eq:bisimulation} guarantees that two different states mapped to the same abstract state share an identical transition.} This condition is crucial for removing the dependence of \emph{interaction complexity} on $\vert \gS \vert$ since it allows estimating the \dquote{abstract transition function}.

\subsection{Algorithm Designs and Theoretical Analysis}

We now extend MB-TAIL to the state abstraction setting. To facilitate later discussion, we introduce some useful notations widely used in the literature \citep{li2006towards, jiang2015abstraction}. In this part, for a function $f$ that operates on the original state space $\gS$, we add a superscript $\phi$ (i.e., $f^{\phi}$) to denote the counterpart that operates on the abstract state space $\Phi$. Inversely, for a function $f^{\phi}$ that operates on the abstract state space, we use $[f^{\phi}]^{M}$ to denote its lifted version, which is defined as $[f^{\phi}]^{M} (s) = f^{\phi} (\phi (s))$. Notice that $[f^{\phi}]^{M}$ is a function over $\gS$.

Now we proceed to present the algorithm MB-TAIL with state abstraction. First, following \cref{algo:framework}, we develop the meta-algorithm with state abstraction, which is outlined in \cref{algo:framework_state_abstraction}. In \cref{algo:framework_state_abstraction}, all learned quantities including transition function, state-action distribution and policy operate in the abstract state space, which is the primary difference from \cref{algo:framework}. At the end of \cref{algo:framework_state_abstraction}, we leverage the lift operator to transform the abstract policy into the original version. 

\begin{algorithm}[htbp]
\caption{Meta-algorithm for AIL with State Abstractions and Unknown Transitions}
\label{algo:framework_state_abstraction}
\begin{algorithmic}[1]
\REQUIRE{Expert demonstrations $\gD$, a set of state abstractions $\{ \phi_h \}_{h=1}^H$.}
\STATE{$\widehat{P}^{\phi} \lar$ Invoke a reward-free exploration method to collect $n$ trajectories and learn an \emph{abstract} transition model. \label{alg_line:reward_free}}
\STATE{$\widetilde{d}_h^{\piE, \phi} \lar $ Estimate the \emph{abstract} expert state-action distribution. \label{alg_line:estimation}}
\STATE{$\widebar{\pi}^{\phi} \lar$ Apply an AIL approach to perform imitation with the expert estimation $\widetilde{d}_h^{\piE, \phi}$ under transition model $\widehat{P}^{\phi}$. \label{alg_line:optimization}}
\ENSURE{Policy $[\widebar{\pi}^{\phi}]^{M}$.}
\end{algorithmic}
\end{algorithm}

In the sequel, we present three main algorithmic designs that appeared in Line \ref{alg_line:reward_free}, Line \ref{alg_line:estimation} and Line \ref{alg_line:optimization} in Algorithm \ref{algo:framework_state_abstraction} in the setting with state abstraction.

\textbf{Reward-free Exploration with State Abstraction.} In this part, we adapt the reward-free exploration method RF-Express to the setting with state abstraction; see \cref{algo:rf_express_state_abstraction} in Appendix \ref{subsec:appendix_rfe_method_state_abstraction}. The main difference is that we learn the abstract transition model and abstract exploration policy. Nevertheless, when interacting with the original environment, we need to transfer the abstract policy $\pi^{\phi, t+1}$ to the lifted version $[\pi^{\phi, t+1}]^{M}$.

\textbf{The Transition-aware Estimator with State Abstraction. }The key idea of the construction of the estimator is similar to that discussed in \cref{subsec:transition_aware_estimator}. However, unlike the original estimator in \eqref{eq:new_estimator_unknown_transition}, the transition-aware estimator with state abstraction is a distribution over the abstract space $\Phi \times \gA$. We present our adaptions to the setting with state abstraction in the following part.

Similar to the procedure presented in \cref{subsec:transition_aware_estimator}, we randomly divide the expert dataset into two equal parts, i.e., $\gD = \gD_1 \cup \gD_1^{c}$ and $\gD_1 \cap \gD_1^{c} = \emptyset$ with $|\gD_1 |=  |\gD_1^{c}| = m / 2$. First, with state abstractions $\{ \phi_h \}_{h=1}^H$, we first apply BC on $\gD_1$ to learn the abstract policy $\pi^{\prime, \phi}$.
\begin{align}
\label{eq:bc_policy_state_abstraction}
\pi^{\prime, \phi}_h(a|x) = \left\{ \begin{array}{ll}
      \frac{n_h^{1}(x, a)}{n_h^{1}(x)}  & \text{if } n_h^{1}(x) > 0  \\
        \frac{1}{|\gA|} & \text{otherwise}  
    \end{array} \right.
\end{align}
Here $n^1_h(x, a) = \sum_{\tr \in \gD_1} \indict \{ \phi_h (\tr_h (\cdot)) = x, \tr_h (a_h) = a \}$ and $n^1_h(x) = \sum_{\tr \in \gD_1} \indict \{ \phi_h (\tr_h (\cdot)) = x \}$. Intuitively, $n^1_h(x, a)$ ($n^1_h(x)$) is the number of abstract-state-action (abstract state) pairs that appeared in $\gD_1$ in step $h$.

Second, we utilize the lifted policy $[\pi^{\prime, \phi}]^M$ to interact with the environment to collect a new dataset $\gD_{\env}^\prime$. Notice that $[\pi^{\prime, \phi}]^M$ is a policy defined in the original state space $\gS$. Finally, we can establish the following estimator with state abstractions $\{ \phi_h \}_{h=1}^H$.
\begin{align}
    \widetilde{d}^{\piE, \phi}_h (x, a) &= \frac{\sum_{\tr_h \in \gD^{\prime}_{\env}} \indict \{ \phi_h (\tr_h (\cdot)) = x, \tr_h (a_h) = a, \tr_h \in \Tr^{\gD_1, \phi}_h   \}}{\vert \gD^{\prime}_{\env} \vert} \nonumber 
    \\
    &+ \frac{\sum_{\tr_h \in \gD_1^c} \indict \{ \phi_h (\tr_h (\cdot)) = x, \tr_h (a_h) = a, \tr_h \not\in  \Tr^{\gD_1, \phi}_h\}}{\labs \gD^{c}_1 \rabs}. \label{eq:new_estimator_unknown_transition_state_abstraction}
\end{align}
Here $\textbf{Tr}^{\gD, \phi}_h = \{ \tr_h = (s_1, a_1, \ldots, s_h, a_h): \phi_{\ell} (s_{\ell}) \in \Phi_{\ell} (\gD), \forall \ell \in [h] \}, \; \Phi_h (\gD) = \{ x \in \Phi: \exists \tr \in \gD, \phi_h (\tr_h (\cdot)) = x\}$. Intuitively, $\Phi_h (\gD)$ is the set of abstract states visited in $\gD$ in time step $h$. $\Tr^{\gD, \phi}_h$ is the set of truncated trajectories of length $h$, along which each abstract state is visited in $\gD$.

\textbf{Gradient-based Optimization.} For Line \ref{alg_line:optimization} in Algorithm \ref{algo:framework_state_abstraction}, we aim to solve the following state-action distribution matching problem.
\begin{align*}
        \min_{\pi^\phi \in \Pi^{\phi}} \sum_{h=1}^H \lnorm \widetilde{d}^{\piE, \phi}_h - d^{\pi^{\phi}, \widehat{P}^{\phi}}_h \rnorm_1,
\end{align*}
Here $\Pi^{\phi}$ is the set of all abstract policies and $d^{\pi^{\phi}, \widehat{P}^{\phi}}_h$ is the abstract state-action distribution of $\pi^{\phi}$ in the model $\widehat{P}^{\phi}$. Notice that this is precisely the optimization problem of projecting $\widetilde{d}^{\piE, \phi}_h$ on the set of all feasible \emph{abstract} state-action distributions. We can still apply \cref{algo:gradient_based_optimization} with inputs of $\widehat{P}^{\phi}$ and $\widetilde{d}^{\piE, \phi}_h$ to solve this optimization problem.

Finally, we combine the above three algorithmic designs under the developed framework (\cref{algo:framework_state_abstraction}), which yields the final algorithm.

\begin{algorithm}[htbp]
\caption{Model-based Transition-aware AIL with State Abstractions}
\label{algo:mbtail-state-abstraction}
\begin{algorithmic}[1]
\REQUIRE{Expert demonstrations $\gD$, and a set of state abstractions $\{ \phi_h \}_{h=1}^H$. }
\STATE{Randomly split $\gD$ into two equal parts: $\gD = \gD_1 \cup \gD_1^{c}$.}
\STATE{Learn an abstract policy $\pi^{\prime, \phi} \in \Pi_{\text{BC}} \lp \gD_{1} \rp$ by BC with $\{ \phi_h \}_{h=1}^H$ and roll out $[\pi^{\prime, \phi}]^{M}$ to obtain dataset $\gD_{\env}^\prime$ with $|\gD_{\env}^\prime| = n^{\prime}$ \label{alg_line:mbtail-state-abstraction-rollout-bc}}.
\STATE{Obtain the abstract estimator $\widetilde{d}_h^{\piE, \phi}$ in \eqref{eq:new_estimator_unknown_transition_state_abstraction} with $\gD$, $\gD_{\env}^\prime$ and $\{ \phi_h \}_{h=1}^H$.}
\STATE{Invoke \cref{algo:rf_express_state_abstraction} to collect $n$ trajectories and learn an abstract empirical transition function $\widehat{P}^{\phi}$.}
\STATE{$\widebar{\pi}^{\phi} \lar$ Apply \cref{algo:gradient_based_optimization} with the estimation $\widetilde{d}_h^{\piE, \phi}$ under transition model $\widehat{P}^{\phi}$.}
\ENSURE{Policy $[\widebar{\pi}^{\phi}]^{M}$.}
\end{algorithmic}
\end{algorithm}

We prove that under \cref{asmp:state_abstraction}, MB-TAIL achieves expert sample and interaction complexities that are independent of $|\gS|$. However, the proof is not straightforward, and the primary challenge is to connect the state-action distributions in the original and abstract MDPs. We provide a detailed discussion of the specialized analysis tools in the Appendix.

\begin{thm}\label{theorem:sample-complexity-unknown-transition-state-abstraction}
Under \cref{asmp:state_abstraction}, fix $\varepsilon \in \lp 0, 1 \rp$ and $\delta \in (0, 1)$; suppose $H \geq 5$. Under the unknown transition setting, consider Algorithm \ref{algo:mbtail-state-abstraction} in Appendix and $[\widebar{\pi}^{\phi}]^{M}$ is output policy. Assume that the RL error $\varepsilon_{\rl} \leq \varepsilon / 2$, the number of iterations $T \gtrsim H^2 |\Phi|  |\gA| / \varepsilon^2$, and the step size $\eta^{(t)} :=  \sqrt{|\Phi||\gA| / (8T)}$. If the number of expert trajectories ($m$), the number of interaction trajectories for estimation ($n^\prime$), and the number of interaction trajectories for reward-free exploration ($n$) satisfy
\begin{align*}
&m \gtrsim  \frac{  |\Phi| H^{3/2}}{\varepsilon} \log\lp\frac{ |\Phi| H}{\delta} \rp, n^{\prime} \gtrsim  \frac{  |\Phi| H^2}{\varepsilon^2} \log \lp \frac{ |\Phi| H }{\delta} \rp, n \gtrsim \frac{ |\Phi| |\gA| H^3}{\varepsilon^2} \lp |\Phi| + \log \lp \frac{ |\Phi| |\gA| H}{\delta \varepsilon} \rp \rp,
\end{align*}
then with probability at least $1-\delta$, we have the imitation gap $V^{\piE} - V^{[\widebar{\pi}^{\phi}]^{M}} \leq \varepsilon $.
\end{thm}

\section{Conclusion}

This paper contributes to the establishment of theoretical foundations for AIL with unknown transitions. We propose a new and general framework that enables AIL to explore and imitate efficiently. As mentioned, AIL methods can have much better theoretical guarantees on structured instances, such as horizon-free bounds suggested in \citep{xu2022understanding}. Thus, we believe that investigating AIL with unknown transitions on structured instances is an interesting and valuable direction for future research.

\section*{Acknowledgment}
Tian Xu would like to thank Zhilong Zhang, Fanming Luo, and Jingcheng Pang for reading the manuscript and providing helpful comments. The work of Yang Yu is supported by National Key Research and Development Program of China (2020AAA0107200), NSFC(61876077), and Collaborative Innovation Center of Novel Software Technology and Industrialization. The work of Zhi-Quan Luo is supported in part by the National Key Research and Development Project under grant 2022YFA1003900, and in part by the Guangdong Provincial Key Laboratory of Big Data Computing.

\bibliographystyle{abbrvnat}
\bibliography{uai_reference.bib}

\newpage
\onecolumn %
\appendix
\section{Notation}
\label{sec:notation}

\begin{table}[H]
\caption{Notations}
\label{table:notations}
\centering
\begin{tabular}{@{}ll@{}}
\toprule
Symbol                     & Meaning \\ \midrule
$\piE$ & the expert policy \\
$V^{\pi, P, r}$ & policy value under the transition model $P$ and reward $r$ \\
$\varepsilon$ & the imitation gap \\
$\delta$ & failure probability \\
$d^{\pi}_h (s)$ & state distribution \\
$d^{\pi}_h (s, a)$ & state-action distribution \\
$\tr = \lp s_1, a_1, \cdots, s_{H}, a_{H} \rp$ & the trajectory \\
$\tr_h = \lp s_1, a_1, \cdots, s_{h}, a_{h} \rp$ & the truncated trajectory \\
$\tr_h(\cdot)$ & the state at time step $h$ in $\tr$ \\
$\tr_h(\cdot, \cdot)$ & the state-action pair at time step $h$ in $\tr$ \\
$\tr (a_h)$ & the action at time step $h$ in $\tr$ \\
$\gD$ & expert dataset \\
$m$   & number of expert trajectories \\
$\widehat{d}_h^{\piE} (s, a)$ & maximum likelihood estimator of $d^{\piE}_h$ in \cref{eq:estimate_by_count}
\\
$\widetilde{d}_h^{\piE} (s, a)$ & transition-aware estimator in \cref{eq:new_estimator_unknown_transition}
\\
$\sP^{\piE}(\tr)$ & probability of the trajectory $\tr$ under the expert policy $\piE$
\\
$\sP^{\piE}(\tr_h)$ & probability of the truncated trajectory $\tr_h$ under the expert policy $\piE$
\\
$\gS_h (\gD)$ & the set of states visited in time step $h$ in dataset $\gD$
\\
$\Tr_h^{\gD}$ & the trajectories along which each state has been visited in $\gD$ up to time step $h$
\\
$\pi^{(t)}$ & the policy obtained in the iteration $t$ \\
$w^{(t)}$ & the reward function learned in the iteration $t$ \\
$\eta^{(t)}$ & the step size in the iteration $t$ \\
$f^{(t)} (w)$ & the objective function in the iteration $t$ in \cref{eq:objective_w} \\
$\widebar{d}_h (s, a)$ & the averaged state-action distribution in \cref{algo:gradient_based_optimization} \\
$\widebar{\pi}$ & the policy derived by the averaged state-action distribution in \cref{algo:gradient_based_optimization} \\
$\Pi_{\text{BC}} \lp \gD_{1} \rp$ & the set of policies which take the expert action on states covered in $\gD_{1}$
\\
$\widehat{P}$ & the empirical transition function
\\
$d^{\pi, \widehat{P}}_h (s, a)$ & the state-action distribution of $\pi$ under the empirical transition function $\widehat{P}$
\\\bottomrule
\end{tabular}%
\end{table}
\section{From Regret Guarantee to PAC Guarantee}
\label{sec:from_regret_to_pac}

\citet{shani2022online} proved a regret guarantee for their OAL algorithm. In particular, \citet{shani2022online} showed that with probability at least $1 - \delta^\prime$, we have 
\begin{align}   \label{eq:oal_regret}
     \sum_{k=1}^{K}  V^{\piE} - V^{\pi_k}  \leq  \widetilde{\gO}\lp \sqrt{H^4 |\gS|^2 |\gA|K} + \sqrt{H^3 |\gS| |\gA| K^2 /m} \rp,
\end{align}
where $\pi^{k}$ is the policy obtained at episode $k$, $K$ is the number of interaction episodes, and $m$ is the number of expert trajectories. We would like to comment that the second term in \eqref{eq:oal_regret} involves the statistical estimation error about the expert policy. Furthermore, this term reduces to $\widetilde{\gO}(\sqrt{H^2 |\gS| K^2 /m})$ under the assumption that the expert policy is deterministic.

To further convert this regret guarantee to the PAC guarantee considered in this paper, we can apply Markov's inequality as suggested by \citep{jin18qlearning}. Concretely, let $\widebar{\pi}$ be the policy that randomly chosen from $\{\pi^{1}, \pi^{2}, \cdots, \pi^{K}\}$ with equal probability, then we have 
\begin{align*}
    \sP \lp V^{\piE} - V^{\widebar{\pi}} \geq \varepsilon \rp \leq \frac{1}{\varepsilon} \expect \ls \frac{1}{K} \sum_{k=1}^{K}  V^{\piE} - V^{\pi_k} \rs \leq \frac{1}{\varepsilon} \lp \widetilde{\gO} \lp \sqrt{ \frac{H^4 |\gS|^2 |\gA|}{K} } + \sqrt{H^2 |\gS| /m}\rp + \delta^\prime H  \rp,
\end{align*}
Therefore, if we set $\delta^{\prime} = \varepsilon \delta / (3H)$, and
\begin{align*}
    K = \widetilde{\gO} \lp \frac{H^4 |\gS|^2 |\gA|}{\varepsilon^2 \delta^2} \rp, \quad m = \widetilde{\gO} \lp \frac{H^2 |\gS|}{\varepsilon^2} \rp,
\end{align*}
we obtain that $\sP ( V^{\piE} - V^{\widebar{\pi}} \geq \varepsilon ) \leq \delta $.

\section{Proof of Results in Section \ref{sec:warm-up}}
\label{sec:proof_of_sec:warm-up}

\subsection{Proof of Lemma \ref{lem:1}}

\begin{proof}
The proof starts with the dual representation of policy value (see \cref{eq:dual_of_policy_value}).
\begin{align*}
    V^{\piE} - V^{\widebar{\pi}} & = \sum_{h=1}^H \sum_{(s, a) \in \gS \times \gA} \lp d^{\piE}_h (s, a) - d^{\widebar{\pi}}_h (s, a) \rp r_h (s, a)
    \\
    &\overset{(a)}{\leq} \sum_{h=1}^H \lnorm d^{\piE}_h - d^{\widebar{\pi}}_h   \rnorm_1
    \\
    &\leq \sum_{h=1}^H \lnorm d^{\piE}_h  - \widetilde{d}^{\piE}_h   \rnorm_1 + \sum_{h=1}^H \lnorm  d^{\widebar{\pi}}_h  - \widetilde{d}^{\piE}_h  \rnorm_1, 
\end{align*}
where inequality $(a)$ is based on the assumption that $r_h (s, a) \in [0, 1]$. For the two terms in RHS, according to \cref{def:estimation} and \cref{def:distribution_matching_error}, we have
\begin{align*}
    \sum_{h=1}^H \lnorm d^{\piE}_h  - \widetilde{d}^{\piE}_h  \rnorm_1 \leq \varepsilon_{\est}, \; \sum_{h=1}^H \lnorm  d^{\widebar{\pi}}_h - \widetilde{d}^{\piE}_h   \rnorm_1 \leq \min_{\pi \in \Pi} \sum_{h=1}^H \lnorm  d^{\pi}_h - \widetilde{d}^{\piE}_h    \rnorm_1 + \varepsilon_{\opt}.  
\end{align*}
With the above two inequalities, we further obtain
\begin{align*}
    V^{\piE} - V^{\widebar{\pi}} &\leq \varepsilon_{\est} + \min_{\pi \in \Pi} \sum_{h=1}^H \lnorm  d^{\pi}_h  - \widetilde{d}^{\piE}_h   \rnorm_1 + \varepsilon_{\opt}
    \\
    &\overset{(a)}{\leq} \varepsilon_{\est} + \sum_{h=1}^H \lnorm  d^{\piE}_h - \widetilde{d}^{\piE}_h    \rnorm_1 + \varepsilon_{\opt}
    \\
    &\leq 2 \varepsilon_{\est} + \varepsilon_{\opt}.  
\end{align*}
Inequality $(a)$ holds since $\piE \in \Pi$. We complete the proof.

\end{proof}
\section{Proof of Results in Section \ref{sec:main_result}}
\label{sec:proof:results_in_sction_main_result}

\subsection{Proof of Proposition \ref{prop:connection}}
\label{subsec:proof_of_proposition_connection}

\begin{proof}
Let $\widetilde{d}^{\piE}_h(s, a)$ be an expert state-action distribution estimator and $\widehat{P}$ be a transition model learned by a reward-free method. Notice that reward-free exploration methods also enable uniform policy evaluation with respect to \emph{any} reward function; see \cref{defn:uniform_policy_evaluation}. That is, with probability at least $1-\delta_{\rfe}$, for any reward function $r$ and policy $\pi$, we have $\vert V^{\pi, P, r} - V^{\pi, \widehat{P}, r} \vert \leq \varepsilon_{\rfe}$. Then we define the following two events.
\begin{align*}
    &E_{\mathrm{EST}}:= \lb \sum_{h=1}^H \lnorm \widetilde{d}^{\piE}_h - d^{\piE}_h  \rnorm_{1} \leq \varepsilon_{\mathrm{EST}} \rb,
    \\
    & E_{\mathrm{RFE}}:= \lb \forall r = (r_1, \ldots, r_H), \forall \pi \in \Pi:  \left\vert V^{\pi, P, r} - V^{\pi, \widehat{P}, r} \right\vert \leq \varepsilon_{\mathrm{RFE}} \rb. 
\end{align*}
According to assumption (a) and (b), we have that $\sP \lp E_{\mathrm{EST}}  \rp \geq 1 - \delta_{\mathrm{EST}}$ and $\sP \lp E_{\mathrm{RFE}}  \rp \geq 1 - \delta_{\mathrm{RFE}}$. Applying union bound yields
\begin{align*}
    \sP \lp E_{\mathrm{EST}}  \cap E_{\mathrm{RFE}} \rp \geq 1 - \delta_{\mathrm{EST}} -  \delta_{\mathrm{RFE}}.
\end{align*}
The following analysis is established on the event $E_{\mathrm{EST}}  \cap E_{\mathrm{RFE}}$. Let $\widebar{\pi}$ be the output of Algorithm \ref{algo:framework}.
\begin{align*}
    \left\vert V^{\piE, P} - V^{\widebar{\pi}, P} \right\vert \leq \left\vert V^{\piE, P} - V^{\widebar{\pi}, \widehat{P}} \right\vert + \left\vert V^{\widebar{\pi}, \widehat{P}} - V^{\widebar{\pi}, P} \right\vert \leq \left\vert V^{\piE, P} - V^{\widebar{\pi}, \widehat{P}} \right\vert + \varepsilon_{\mathrm{RFE}}. 
\end{align*}
The last inequality follows the event $E_{\mathrm{RFE}}$. Then we consider the error $\vert V^{\piE, P} - V^{\widebar{\pi}, \widehat{P}} \vert$. From the dual form of the policy value in \cref{eq:dual_of_policy_value}, we have that
\begin{align*}
    \left\vert V^{\piE, P} - V^{\widebar{\pi}, \widehat{P}} \right\vert &= \left\vert \sum_{h=1}^H \sum_{(s, a) \in \gS \times \gA} \lp d^{\piE, P}_h (s, a) - d^{\widebar{\pi}, \widehat{P}}_h (s, a) \rp r_h (s, a)  \right\vert \leq \sum_{h=1}^H \lnorm d^{\piE, P}_h - d^{\widebar{\pi}, \widehat{P}}_h   \rnorm_1,
\end{align*}
where $d^{\widebar{\pi}, \widehat{P}}_h (s, a)$ is the state-action distribution of the policy $\widebar{\pi}$ under the transition model $\widehat{P}$. Then we get that
\begin{align*}
    \sum_{h=1}^H \lnorm d^{\piE, P}_h - d^{\widebar{\pi}, \widehat{P}}_h   \rnorm_1 & \leq \sum_{h=1}^H \lnorm d^{\piE, P}_h - \widetilde{d}^{\piE}_h   \rnorm_1 + \sum_{h=1}^H \lnorm \widetilde{d}^{\piE}_h - d^{\widebar{\pi}, \widehat{P}}_h   \rnorm_1
    \\
    &\leq \varepsilon_{\mathrm{EST}} + \sum_{h=1}^H \lnorm \widetilde{d}^{\piE}_h - d^{\widebar{\pi}, \widehat{P}}_h   \rnorm_1. 
\end{align*}
The last inequality follows the event $E_{\mathrm{EST}}$. Combining the above three inequalities yields
\begin{align*}
    \left\vert V^{\piE, P} - V^{\widebar{\pi}, P} \right\vert \leq \sum_{h=1}^H \lnorm \widetilde{d}^{\piE}_h - d^{\widebar{\pi}, \widehat{P}}_h   \rnorm_1 + \varepsilon_{\mathrm{EST}} +  \varepsilon_{\mathrm{RFE}}.
\end{align*}
According to assumption (c), with the estimator $\widetilde{d}^{\piE}_h (s, a)$ and transition model $\widehat{P}$, algorithm C solves the optimization problem in \cref{eq:ail_with_model} up to an error $\varepsilon_{\opt}$ and $\widebar{\pi}$ is the output of the algorithm C. Formally,
\begin{align*}
     \sum_{h=1}^H \lnorm \widetilde{d}^{\piE}_h - d^{\widebar{\pi}, \widehat{P}}_h   \rnorm_1 \leq \min_{\pi \in \Pi} \sum_{h=1}^H \lnorm \widetilde{d}^{\piE}_h - d^{\pi, \widehat{P}}_h   \rnorm_1 + \varepsilon_{\opt}.
\end{align*}
Then we get that
\begin{align*}
    \left\vert V^{\piE, P} - V^{\widebar{\pi}, P} \right\vert &\leq \sum_{h=1}^H \lnorm \widetilde{d}^{\piE}_h - d^{\widebar{\pi}, \widehat{P}}_h   \rnorm_1 + \varepsilon_{\mathrm{EST}} +  \varepsilon_{\mathrm{RFE}}
    \\
    &\leq \min_{\pi \in \Pi} \sum_{h=1}^H \lnorm \widetilde{d}^{\piE}_h - d^{\pi, \widehat{P}}_h  \rnorm_1 +  \varepsilon_{\opt} + \varepsilon_{\mathrm{EST}} +  \varepsilon_{\mathrm{RFE}}
    \\
    &\overset{(a)}{\leq} \sum_{h=1}^H \lnorm \widetilde{d}^{\piE}_h - d^{\piE, \widehat{P}}_h   \rnorm_1 +  \varepsilon_{\opt} + \varepsilon_{\mathrm{EST}} +  \varepsilon_{\mathrm{RFE}}
    \\
    &\leq \sum_{h=1}^H \lnorm \widetilde{d}^{\piE}_h - d^{\piE, P}_h   \rnorm_1 + \sum_{h=1}^H \lnorm d^{\piE, P}_h - d^{\piE, \widehat{P}}_h   \rnorm_1  +  \varepsilon_{\opt} + \varepsilon_{\mathrm{EST}} +  \varepsilon_{\mathrm{RFE}}
    \\
    &\overset{(b)}{\leq} \sum_{h=1}^H \lnorm d^{\piE, P}_h - d^{\piE, \widehat{P}}_h   \rnorm_1  +  \varepsilon_{\opt} + 2\varepsilon_{\mathrm{EST}} +  \varepsilon_{\mathrm{RFE}},
\end{align*}
where inequality $(a)$ holds since $\piE \in \Pi$ and inequality $(b)$ follows the event $E_{\mathrm{EST}}$. With the dual representation of $\ell_1$-norm, we have that
\begin{align*}
    \sum_{h=1}^H \lnorm d^{\piE, P}_h - d^{\piE, \widehat{P}}_h \rnorm_1 &= \max_{w \in \gW} \sum_{h=1}^H \sum_{(s, a) \in \gS \times \gA} w_h (s, a) \lp d^{\piE, P}_h (s, a) - d^{\piE, \widehat{P}}_h (s, a)  \rp
    \\
    &= \max_{w \in \gW} \sum_{h=1}^H V^{\piE, P, w} - V^{\piE, \widehat{P}, w} \leq \varepsilon_{\mathrm{RFE}},
\end{align*}
where $\gW = \{w = (w_1, \ldots, w_H): w_h \in \reals^{|\gS| \times |\gA|}, \|w_h \|_{\infty} \leq 1 \}$, $V^{\piE, \widehat{P}, w}$ is the value of policy $\piE$ with the transition model $\widehat{P}$ and reward function $w$. The last inequality follows the event $E_{\mathrm{RFE}}$. Then we prove that
\begin{align*}
    \left\vert V^{\piE, P} - V^{\widebar{\pi}, P} \right\vert \leq  2\varepsilon_{\mathrm{EST}} + 2 \varepsilon_{\mathrm{RFE}} + \varepsilon_{\opt}. 
\end{align*}
\end{proof}

\subsection{Reward-free Exploration Method}

In this part, we present the RF-Express algorithm in \citep{menard20fast-active-learning} with our notations. Please see \cref{algo:rf_express}.

\begin{algorithm}[htbp]
\caption{RF-Express}
\label{algo:rf_express}
\begin{algorithmic}[1]
\REQUIRE{Failure probability $\delta$, function $\beta(n, \delta) = \log (3 \vert \gS \vert \vert \gA \vert H / \delta)+ \vert \gS \vert \log (8 e(n+1))$.}
\FOR{$t = 0, 1, 2, \cdots$}
\STATE{Update the counter and the empirical transition model:}
\begin{align*}
    & n^{t}_h (s, a) = \sum_{i=1}^{t} \indict \{ s^{i}_h = s, a^i_h = a \}, \; n^{t}_h (s, a, s^\prime) = \sum_{i=1}^{t} \indict \{ s^{i}_h = s, a^i_h = a, s^{i}_{h+1} = s^\prime  \},
    \\
    & \widehat{P}^{t}_h (s^\prime|s, a) = \frac{n^{t}_h (s, a, s^\prime)}{n^{t}_h (s, a)}, \; \text{if $n^{t}_h (s, a) > 0$ and } \widehat{P}^{t}_h (s^\prime|s, a) = \frac{1}{\vert \gS \vert}, \; \forall s^\prime \in \gS \text{ otherwise}. 
\end{align*}
\STATE{Define $W^{t}_{H+1} (s, a) = 0, \; \forall (s, a) \in \gS \times \gA$.}
\FOR{$h = H, H-1, \cdots, 1$}
\STATE{$W^{t}_h (s, a) = \min \left(H, 15 H^{2} \frac{\beta\left(n_{h}^{t } (s, a), \delta\right)}{n_{h}^{t } (s, a)}+\left(1+\frac{1}{H}\right) \sum_{s^{\prime} \in \gS} \widehat{P}_{h}^{t} (s^{\prime} | s, a ) \max_{a^{\prime}} W_{h+1}^{t }\left(s^{\prime}, a^{\prime}\right)\right)$.}
\ENDFOR
\STATE{Derive the greedy policy: $\pi_{h}^{t+1}(s)=\argmax_{a \in \mathcal{A}} W_{h}^{t }(s, a), \forall s \in \gS, \forall h \in[H]$.}
\IF{$3 e \sqrt{ W_{1}^{t} (s_1, \pi_{1}^{t+1, }(s_1)) }+ W_{1}^{t} (s_1, \pi_{1}^{t+1}(s_1)) \leq \varepsilon / 2$}
\BREAK
\ENDIF
\STATE{Rollout $\pi^{t+1}$ to collect a trajectory $\tau^{t+1} = (s^t_1, a^t_1, s^t_2, a^t_2, \cdots, s^t_H, a^t_H)$.}
\ENDFOR
\ENSURE{Transition model $\widehat{P}^{t}$.}
\end{algorithmic}
\end{algorithm}

\subsection{Proof of Lemma \ref{lemma:sample_complexity_of_new_estimator_unknown_transition}}

Prior to proving \cref{lemma:sample_complexity_of_new_estimator_unknown_transition}, we first prove that the estimator shown in \eqref{eq:new_estimator_unknown_transition} is an unbiased estimation. We consider the decomposition of $d_h^{\piE} (s, a)$.
\begin{align} 
d_h^{\piE}(s, a) &= {\sum_{\tr_h \in \Tr_h^{\gD_1}} \sP^{\piE}(\tr_h) \indict\lb \tr_h(\cdot, \cdot) = (s, a) \rb} + {\sum_{\tr_h \notin \Tr_h^{\gD_1}} \sP^{\piE}(\tr_h) \indict\lb \tr_h(\cdot, \cdot) = (s, a) \rb}  \nonumber 
\\
&= {\sum_{\tr_h \in \Tr_h^{\gD_1}} \sP^{\pi^\prime}(\tr_h) \indict\lb \tr_h(\cdot, \cdot) = (s, a) \rb} + {\sum_{\tr_h \notin \Tr_h^{\gD_1}} \sP^{\piE}(\tr_h) \indict\lb \tr_h(\cdot, \cdot) = (s, a) \rb},  \label{eq:proof_1}
\end{align}
where $\pi^{\prime} \in \Pi_{\text{BC}} \lp \gD_1 \rp$ and the last equality follows Lemma \ref{lemma:unknown-transition-unbiased-estimation}.

\begin{lem} \label{lemma:unknown-transition-unbiased-estimation}
We define $\Pi_{\text{BC}} \lp \gD_1 \rp$ as the set of policies, each of which takes expert action on states contained in $\gD_{1}$. For each $\pi \in \Pi_{\text{BC}} \lp \gD_{1} \rp$, $\forall h \in [H]$ and $(s, a) \in \gS \times \gA$, we have
\begin{align*}
    \sum_{\tr_h \in \Tr_h^{\gD_1}} \sP^{\piE}(\tr_h) \indict\lb \tr_h(\cdot, \cdot) = (s, a) \rb 
    =\sum_{\tr_h \in \Tr_h^{\gD_1}} \sP^{\pi}(\tr_h) \indict\lb \tr_h(\cdot, \cdot) = (s, a) \rb.
\end{align*}
\end{lem}

\begin{proof}
The proof is based on the fact that any $\pi \in \Pi_{\bc}(\gD_1)$ takes the same action with the expert on trajectories in $\Tr_h^{\gD_1}$. More concretely, for any $\tr_h \in \Tr_h^{\gD_1}$, we have 
\begin{align*}
    &\quad \sP^{\piE}(\tr_h) \\
    &= \rho (\tr_h(s_1)) \piE_1 \lp \tr_h(a_1)| \tr(s_1) \rp \prod_{\ell=1}^{h-1}  P_{\ell} \lp \tr_h(s_{\ell+1}) | \tr_h(s_{\ell}), \tr_h(a_{\ell}) \rp \piE_{\ell+1} \lp \tr_h(a_{\ell+1}) | \tr_h(s_{\ell+1}) \rp
    \\
    &= \rho (\tr_h(s_1)) \pi_1 \lp \tr_h(a_1)| \tr(s_1) \rp \prod_{\ell=1}^{h-1}  P_{\ell} \lp \tr_h(s_{\ell+1}) | \tr_h(s_{\ell}), \tr_h(a_{\ell}) \rp \pi_{\ell+1} \lp \tr_h(a_{\ell+1}) | \tr_h(s_{\ell+1}) \rp
    \\
    &= \sP^{\pi}(\tr_h),
\end{align*}
which completes the proof.
\end{proof}

Now we proceed to prove \cref{lemma:sample_complexity_of_new_estimator_unknown_transition}.
\begin{proof}[Proof of \cref{lemma:sample_complexity_of_new_estimator_unknown_transition}]
    
We aim to upper bound the estimation error $\sum_{h=1}^H \Vert \widetilde{d}^{\piE}_h - d^{\piE}_h  \Vert_{1}$. Recall the definition of the estimator $\widetilde{d}^{\piE}_h(s, a)$ in \cref{eq:new_estimator_unknown_transition}:
\begin{align*}
    \widetilde{d}_h^{\piE} (s, a) := \frac{\sum_{\tr_h \in \gD^\prime_{\env}} \indict \lb \tr_h (\cdot, \cdot) = (s, a), \tr_h \in \Tr_h^{\gD_1} \rb }{ | \gD^\prime_{\env} |} + \frac{\sum_{\tr_h \in \gD_1^{c}} \indict \lb \tr_h (\cdot, \cdot) = (s, a), \tr_h \notin \Tr_h^{\gD_1} \rb}{| \gD_1^{c} |}.
\end{align*}
Using \cref{eq:proof_1}, for any $h \in [H]$ and $(s, a) \in \gS \times \gA$, we have
\begin{align*}
    &\quad \labs \widetilde{d}^{\piE}_h(s, a) - d^{\piE}_h (s, a)  \rabs
    \\
    &\leq \labs \frac{\sum_{\tr_h \in \gD^\prime_{\env}} \indict \lb \tr_h (\cdot, \cdot) = (s, a), \tr_h \in \Tr_h^{\gD_1} \rb }{ | \gD^\prime_{\env} |} - \sum_{\tr_h \in \Tr_h^{\gD_1}} \sP^{\pi^\prime}(\tr_h) \indict\lb  \tr_h (\cdot, \cdot) = (s, a)  \rb  \rabs 
    \\
    & + \labs \frac{\sum_{\tr_h \in \gD_1^{c}} \indict \lb \tr_h (\cdot, \cdot) = (s, a), \tr_h \notin \Tr_h^{\gD_1} \rb}{| \gD_1^{c} |} - \sum_{\tr_h \notin \Tr_h^{\gD_1}} \sP^{\piE}(\tr_h) \indict\lb  \tr_h (\cdot, \cdot) = (s, a)  \rb  \rabs .
\end{align*}
Thus, we can upper bound the estimation error.
\begin{align*}
    &\quad \sum_{h=1}^H \lnorm \widetilde{d}^{\piE}_h - d^{\piE}_h  \rnorm_{1}
    \\
    &\leq \underbrace{\sum_{h=1}^H \sum_{(s, a) \in \gS \times \gA} \labs \frac{\sum_{\tr_h \in \gD^\prime_{\env}} \indict \lb \tr_h (\cdot, \cdot) = (s, a), \tr_h \in \Tr_h^{\gD_1} \rb }{ | \gD^\prime_{\env} |} - \sum_{\tr_h \in \Tr_h^{\gD_1}} \sP^{\pi^{\prime}}(\tr_h) \indict\lb  \tr_h (\cdot, \cdot) = (s, a)  \rb  \rabs}_{\text{Error A}}
    \\
    &+ \underbrace{\sum_{h=1}^H \sum_{(s, a) \in \gS \times \gA} \labs \frac{\sum_{\tr_h \in \gD_1^{c}} \indict \lb \tr_h (\cdot, \cdot) = (s, a), \tr_h \notin \Tr_h^{\gD_1} \rb}{| \gD_1^{c} |} - \sum_{\tr_h \notin \Tr_h^{\gD_1}} \sP^{\piE}(\tr_h) \indict\lb  \tr_h (\cdot, \cdot) = (s, a)  \rb  \rabs}_{\text{Error B}}. 
\end{align*}
We first analyze the term $\text{Error A}$. Trajectories in $\gD^{\prime}_{\env}$ are collected by $\pi^{\prime}$ via interacting with the environment. Thus, we have the estimator in $\text{Error A}$ is unbiased, i.e., for any $(s, a) \in \gS \times \gA$ and $h \in [H]$,
\begin{align*}
    \expect_{}\ls  \frac{\sum_{\tr_h \in \gD^\prime_{\env}} \indict \lb \tr_h (\cdot, \cdot) = (s, a), \tr_h \in \Tr_h^{\gD_1} \rb }{ | \gD^\prime_{\env} |}  \rs =  \sum_{\tr_h \in \Tr_h^{\gD_1}} \sP^{\pi^{\prime}}(\tr_h) \indict\lb  \tr_h (\cdot, \cdot) = (s, a)  \rb,
\end{align*}
where the expectation is taken over the randomness of collecting $\gD_{\env}^{\prime}$. The above equality holds because the stochastic processes on the both sides are induced by $\pi^{\prime}$. Then we leverage Chernoff's bound to upper bound \text{Error} A.
\begin{lem}[Chernoff's bound \citep{vershynin2018high}]   \label{lemma:chernoff_bound}
Let $\widebar{X} = {1}/{n} \cdot \sum_{i=1}^{n} X_i$, where $X_i$ is a Bernoulli random variable with $\sP(X_i = 1) = p_i$ and $\sP(X_i = 0) = 1 - p_i$, for $i \in [n]$. Furthermore, assume these random variables are independent. Let $\mu = \expect[\widebar{X}] = {1}/{n} \cdot \sum_{i=1}^{n} p_i$. Then for $0 < t \leq 1$, 
\begin{align*}
    \sP\lp  \labs  \widebar{X} - \mu \rabs  \geq t \mu  \rp \leq 2 \exp\lp -\frac{\mu n t^2}{3}  \rp.
\end{align*}
\end{lem}
First, for each $s \in \gS$ and $h \in [H]$, for any non-expert action $a \not= \piE_h (s)$, we have that
\begin{align*}
     \sum_{\tr_h \in \Tr_h^{\gD_1}} \sP^{\pi^{\prime}}(\tr_h) \indict\lb  \tr_h (\cdot, \cdot) = (s, a)  \rb = 0.
\end{align*}
This is because on the trajectory $\tr_h \in \Tr^{\gD_1}_h$, the state $s$ in time step $h$ is covered in $\gD_1$. As a result, the BC policy $\pi^\prime$ learned from $\gD_1$ must take the expert action $\piE_h (s)$ on such a state and thus $\sP^{\pi^{\prime}}(\tr_h) \indict\lb  \tr_h (\cdot, \cdot) = (s, a)  \rb = 0$. Second, since the estimator of
\begin{align*}
    \frac{\sum_{\tr_h \in \gD^\prime_{\env}} \indict \lb \tr_h (\cdot, \cdot) = (s, a), \tr_h \in \Tr_h^{\gD_1} \rb }{ | \gD^\prime_{\env} |}
\end{align*}
is an unbiased estimator and is non-negative almost surely. Therefore, for each $s \in \gS$ and $h \in [H]$, for any non-expert action $a \not= \piE_h (s)$, with probability of $1$,  
\begin{align*}
    \frac{\sum_{\tr_h \in \gD^\prime_{\env}} \indict \lb \tr_h (\cdot, \cdot) = (s, a), \tr_h \in \Tr_h^{\gD_1} \rb }{ | \gD^\prime_{\env} |} = 0.
\end{align*}
Based on the above two claims, we have that
\begin{align*}
    \text{Error A} = \sum_{h=1}^H \sum_{s \in \gS} \labs \frac{\sum_{\tr_h \in \gD^\prime_{\env}} \indict \lb \tr_h (\cdot, \cdot) = (s, \piE_h (s)), \tr_h \in \Tr_h^{\gD_1} \rb }{ | \gD^\prime_{\env} |} - \sum_{\tr_h \in \Tr_h^{\gD_1}} \sP^{\pi^{\prime}}(\tr_h) \indict\lb  \tr_h (\cdot, \cdot) = (s, \piE_h (s))  \rb  \rabs. 
\end{align*}

Let ${E^\prime}^{s}_h$ be the event that $\tr_h \in \gD_{\env}^{\prime}$ agrees with expert policy at state $s$ at time step $h$ and also appears in $\Tr_h^{\gD_1}$. Formally, 
\begin{align*}
    {E^\prime}_h^{s} = \indict\{\tr_h (\cdot, \cdot) = (s, \piE_h (s)) \cap \tr_h \in \textbf{Tr}_h^{\gD_1}\}.
\end{align*}

By Lemma \ref{lemma:chernoff_bound}, for each $s \in \gS$ and $h \in [H]$, with probability at least $1 - \frac{\delta}{2 |\gS| H}$ over the randomness of $\gD_{\env}^\prime$, we have
\begin{align*}
    &\quad \labs \frac{\sum_{\tr_h \in \gD^\prime_{\env}} \indict \lb \tr_h (\cdot, \cdot) = (s, \piE_h(s)), \tr_h \in \Tr_h^{\gD_1} \rb }{ | \gD^\prime_{\env} |}  - \sum_{\tr_h \in \Tr_h^{\gD_1}} \sP^{\pi^{\prime}}(\tr_h) \indict\lb  \tr_h (\cdot, \cdot) = (s, \piE_h (s))  \rb  \rabs
    \\
    &\leq \sqrt{ \sP^{\pi^{\prime}} \lp {E^\prime}^{s}_h  \rp  \frac{3 \log \lp 4 |\gS| H / \delta \rp}{n^\prime}}.
\end{align*}
By union bound, with probability at least $1-\frac{\delta}{2}$ over the randomness of $\gD^\prime_{\env}$, we have
\begin{align*}
    \text{Error A} &\leq  \sum_{h=1}^H \sum_{s \in \gS} \sqrt{ \sP^{\pi^{\prime}} \lp {E^\prime}^{s}_h  \rp  \frac{3 \log \lp 4 |\gS| H / \delta \rp}{n^\prime}}
    \\
    &\leq \sum_{h=1}^H \sqrt{|\gS|} \sqrt{\sum_{s \in \gS} \sP^{\pi^{\prime}} \lp {E^\prime}^{s}_h  \rp  \frac{3 \log \lp 4 |\gS| H / \delta \rp}{n^\prime} }
\end{align*}
The last inequality follows the Cauchy-Schwartz inequality. It remains to upper bound $\sum_{s \in \gS}  \sP^{\piE}({E^\prime}_{h}^{s})$ for all $h \in [H]$. To this end, we define the event ${G^\prime}_h^{\gD_1}$ that policy $\pi^{\prime}$ visits states covered in $\gD_1$ up to time step $h$. Formally, ${G^\prime}_h^{\gD_1} = \indict\{ \forall h^{\prime} \leq h,  s_{h^{\prime}} \in \gS_{h^{\prime}} (\gD_1) \}$, where $\gS_{h}(\gD_1)$ is the set of states in $\gD_1$ at time step $h$, where $s_h^{\prime}$ comes from $\tr_h \in \gD_{\env}^{\prime}$. Then, for all $h \in [H]$, we have 
\begin{align*}
    \sum_{s \in \gS} \sP^{\pi^{\prime}} \lp {E^\prime}_h^{s}  \rp = \sP^{\pi^{\prime}}({G^\prime}_h^{\gD_1}) \leq \sP({G^\prime}_1^{\gD_1}).
\end{align*}
The last inequality holds since ${G^\prime}_h^{\gD_1} \subseteq {G^\prime}_1^{\gD_1}$ for all $h \in [H]$. Then we have that
\begin{align*}
    \text{Error A} \leq H \sqrt{\frac{3 |\gS| \log \lp 4 |\gS| H / \delta \rp}{n^\prime}}.
\end{align*}
When the interaction complexity satisfies that $n^\prime \gtrsim \frac{| \gS | H^{2}}{\varepsilon^2} \log\lp  \frac{|\gS| H}{\delta} \rp$, with probability at least $1-\frac{\delta}{2}$ over the randomness of $\gD^\prime$, we have $\text{Error A} \leq \frac{\varepsilon}{2}$.

For the term $\text{Error B}$, we utilize \citep[Lemma A.11]{rajaraman2020fundamental}. When the expert sample complexity satisfies that $m \gtrsim \frac{|\gS| H^{3/2}}{\varepsilon} \log \lp \frac{|\gS| H}{\delta} \rp$, with probability at least $1-\frac{\delta}{2}$ over the randomness of $\gD$, we have $\text{Error B} \leq \frac{\varepsilon}{2}$. Applying union bound finishes the proof.

\end{proof}

\subsection{Proof of Lemma \ref{lemma:approximate-minimax}}
Before we prove Lemma \ref{lemma:approximate-minimax}, we first state the following key lemma.
\begin{lem}\label{lemma:regret_of_ogd}
Consider Algorithm \ref{algo:gradient_based_optimization}, we have
\begin{align*}
    \sum_{t=1}^T f^{(t)} \lp w^{(t)} \rp - \min_{w \in \gW} \sum_{t=1}^T f^{(t)} (w) \leq 2H \sqrt{2 |\gS| |\gA| T},
\end{align*}
where $f^{(t)}(w) = \sum_{h=1}^{H} \sum_{(s, a) \in \gS \times \gA} w_h(s, a) ( d^{\pi^{(t)}, \widehat{P}}_h (s, a) - \widetilde{d}^{\piE}_h (s, a) )$.
\end{lem}
\begin{proof}
Lemma \ref{lemma:regret_of_ogd} is a direct consequence of the regret bound of online gradient descent \citep{shalev12online-learning}. To apply such a regret bound, we need to verify that 1) the iterate norm $\lnorm w \rnorm_2$ has an upper bound; 2) the gradient norm $\Vert \nabla_{w}  f^{(t)}(w) \Vert_2$ also has an upper bound. The first point is easy to show, i.e., $\lnorm w \rnorm_2 \leq \sqrt{H |\gS| |\gA|}$ by the condition that $w \in \gW = \{ w = (w_1, \ldots, w_H): \Vert w_h \Vert_{\infty} \leq 1, \; \forall h \in [H] \}$. For the second point, let $\widetilde{d}^{1}_h$ and $\widetilde{d}^{2}_h$ be the first and the second part in $\widetilde{d}^{\piE}_h$ defined in \eqref{eq:new_estimator_unknown_transition}. Then, 
\begin{align*}
    \lnorm \nabla_{w} f^{(t)} (w) \rnorm_{2} &= \sqrt{\sum_{h=1}^H \sum_{(s, a) \in \gS \times \gA} \lp d^{\pi^{(t)}, \widehat{P}}_h (s, a) - \widetilde{d}^{\piE}_h (s, a) \rp^2 }
    \\
    &= \sqrt{\sum_{h=1}^H \sum_{(s, a) \in \gS \times \gA} \lp d^{\pi^{(t)}, \widehat{P}}_h (s, a) - \widetilde{d}^{1}_h(s, a) - \widetilde{d}^{2}_h(s, a) \rp^2 } 
    \\
    &\leq \sqrt{\sum_{h=1}^H  3 \sum_{(s, a) \in \gS \times \gA}   \lp d^{\pi^{(t)}, \widehat{P}}_h (s, a) \rp^2 + \lp \widetilde{d}^{1}_h(s, a) \rp^2 + \lp \widetilde{d}^{2}_h(s, a)  \rp^2 }
    \\
     &\leq \sqrt{\sum_{h=1}^H  3  \lp \lnorm d^{\pi^{(t)}, \widehat{P}}_h \rnorm_1 + \lnorm \widetilde{d}^{1}_h \rnorm_1 + \lnorm \widetilde{d}^{2}_h \rnorm_1  \rp }
    \\
    &\leq 2\sqrt{ H },
\end{align*}
where the first inequality follows $(a+b+c)^2 \leq 3(a^2+b^2+c^2)$ and the second inequality is based on that $ x ^2 \leq \vert x \vert$ if $0 \leq x \leq 1$.

Invoking Corollary 2.7 in \citep{shalev12online-learning} with $B = \sqrt{H |\gS| |\gA|}$ and $L = 2 \sqrt{H}$ finishes the proof. 
\end{proof}

\begin{proof}[Proof of \cref{lemma:approximate-minimax}]
    
 With the dual representation of $\ell_1$-norm, we have
\begin{align*}
    \min_{\pi \in \Pi} \sum_{h=1}^H \lnorm d^{\pi, \widehat{P}}_h - \widetilde{d}^{\piE}_h \rnorm_{1} = \min_{\pi \in \Pi} \max_{w \in \gW} \sum_{h=1}^H \sum_{(s, a) \in \gS \times \gA} w_{h} (s, a) \lp \widetilde{d}^{\piE}_h(s, a) - d^{\pi, \widehat{P}}_h (s, a) \rp. 
\end{align*}
Since the above objective is linear w.r.t both $w$ and $d^\pi_h$, invoking the minimax theorem \citep{bertsekas2016nonlinear} yields
\begin{align*}
    &\quad \min_{\pi \in \Pi} \max_{w \in \gW} \sum_{h=1}^H \sum_{(s, a) \in \gS \times \gA} w_{h} (s, a) \lp \widetilde{d}^{\piE}_h(s, a) - d^{\pi, \widehat{P}}_h (s, a) \rp
    \\
    &= \max_{w \in \gW} \min_{\pi \in \Pi} \sum_{h=1}^H \sum_{(s, a) \in \gS \times \gA} w_{h} (s, a) \lp \widetilde{d}^{\piE}_h(s, a) - d^{\pi, \widehat{P}}_h (s, a) \rp
    \\
    &= - \min_{w \in \gW} \max_{\pi \in \Pi} \sum_{h=1}^H \sum_{(s, a) \in \gS \times \gA} w_h (s, a) \lp d^{\pi, \widehat{P}}_h (s, a) -  \widetilde{d}^{\piE}_h(s, a) \rp, 
\end{align*}
where the last step follows the property that for a function $f$, $- \max_{x} f(x) = \min_{x} - f(x)$. Therefore, we have
\begin{align} \label{eq:l1_dual_representation}
    \min_{\pi \in \Pi} \sum_{h=1}^H \lnorm d^{\pi, \widehat{P}}_h  - \widetilde{d}^{\piE}_h \rnorm_{1} = - \min_{w \in \gW} \max_{\pi \in \Pi} \sum_{h=1}^H \sum_{(s, a) \in \gS \times \gA} w_h (s, a) \lp d^{\pi, \widehat{P}}_h (s, a) -  \widetilde{d}^{\piE}_h(s, a) \rp.
\end{align}
Then we consider the term $\min_{w \in \gW} \max_{\pi \in \Pi} \sum_{h=1}^H \sum_{(s, a) \in \gS \times \gA} w_h (s, a) \lp d^{\pi, \widehat{P}}_h (s, a) -  \widetilde{d}^{\piE}_h(s, a) \rp$.
\begin{align*}
    &\quad \min_{w \in \gW} \max_{\pi \in \Pi} \sum_{h=1}^H \sum_{(s, a) \in \gS \times \gA} w_h (s, a) \lp d^{\pi, \widehat{P}}_h (s, a) -  \widetilde{d}^{\piE}_h(s, a) \rp
    \\
    &\leq \max_{\pi \in \Pi} \sum_{h=1}^H \sum_{(s, a) \in \gS \times \gA} \lp \frac{1}{T} \sum_{t=1}^T w^{(t)}_h (s, a) \rp \lp d^{\pi, \widehat{P}}_h (s, a) -  \widetilde{d}^{\piE}_h(s, a) \rp
    \\
    &\leq \frac{1}{T} \sum_{t=1}^T \max_{\pi \in \Pi} \sum_{h=1}^H \sum_{(s, a) \in \gS \times \gA} w^{(t)}_h (s, a) \lp d^{\pi, \widehat{P}}_h (s, a) -  \widetilde{d}^{\piE}_h(s, a) \rp. 
\end{align*}
At iteration $t$, $\pi^{(t)}$ is the approximately optimal policy regarding reward function $w^{(t)}$ with an optimization error of $\varepsilon_{\rl}$. Then we obtain that
\begin{align*}
    &\quad \frac{1}{T} \sum_{t=1}^T \max_{\pi \in \Pi} \sum_{h=1}^H \sum_{(s, a) \in \gS \times \gA} w^{(t)}_h (s, a) \lp d^{\pi, \widehat{P}}_h (s, a) -  \widetilde{d}^{\piE}_h(s, a) \rp
    \\
    &\leq \frac{1}{T} \sum_{t=1}^T \sum_{h=1}^H \sum_{(s, a) \in \gS \times \gA} w^{(t)}_h (s, a) \lp d^{\pi^{(t)}, \widehat{P}}_h (s, a) -  \widetilde{d}^{\piE}_h(s, a) \rp + \varepsilon_{\rl}.
\end{align*}
Applying Lemma \ref{lemma:regret_of_ogd} yields that
\begin{align*}
    &\quad \frac{1}{T} \sum_{t=1}^T \sum_{h=1}^H \sum_{(s, a) \in \gS \times \gA} w^{(t)}_h (s, a) \lp d^{\pi^{(t)}, \widehat{P}}_h (s, a) -  \widetilde{d}^{\piE}_h(s, a) \rp
    \\
    & \leq \min_{w \in \gW} \frac{1}{T} \sum_{t=1}^T \sum_{h=1}^H \sum_{(s, a) \in \gS \times \gA} w_h (s, a) \lp d^{\pi^{(t)}, \widehat{P}}_h (s, a) -  \widetilde{d}^{\piE}_h(s, a) \rp + 2H \sqrt{ \frac{2 |\gS| |\gA|}{T} }
    \\
    &= \min_{w \in \gW}  \sum_{h=1}^H \sum_{(s, a) \in \gS \times \gA} w_h (s, a) \lp \frac{1}{T} \sum_{t=1}^T d^{\pi^{(t)}, \widehat{P}}_h (s, a) -  \widetilde{d}^{\piE}_h(s, a) \rp + 2H \sqrt{ \frac{2 |\gS| |\gA|}{T} }
    \\
    &= \min_{w \in \gW}  \sum_{h=1}^H \sum_{(s, a) \in \gS \times \gA} w_h (s, a) \lp d^{\widebar{\pi}, \widehat{P}}_h (s, a) -  \widetilde{d}^{\piE}_h(s, a) \rp + 2H \sqrt{ \frac{2 |\gS| |\gA|}{T} }.
\end{align*}
Note that $\widebar{\pi}$ is induced by the mean state-action distribution, i.e., $\widebar{\pi}_h (a|s) = \widebar{P}_h(s, a) / \sum_{a} \widebar{P}_h(s, a)$, where $\widebar{P}_h (s, a) = \frac{1}{T} \sum_{t=1}^T d^{\pi^{(t)}, \widehat{P}}_h (s, a)$. Based on Proposition 3.1 in \citep{ho2016gail}, we have that $d^{\widebar{\pi}, \widehat{P}}_h (s, a) = \widebar{P}_h (s, a)$, and hence the last equation holds. Combined with \cref{eq:l1_dual_representation}, we have that
\begin{align*}
    &\quad \min_{\pi \in \Pi} \sum_{h=1}^H \lnorm d^{\pi, \widehat{P}}_h - \widetilde{d}^{\piE}_h \rnorm_{1}
    \\
    &\geq - \min_{w \in \gW}  \sum_{h=1}^H \sum_{(s, a) \in \gS \times \gA} w_h (s, a) \lp d^{\widebar{\pi}, \widehat{P}}_h (s, a) -  \widetilde{d}^{\piE}_h(s, a) \rp - 2H \sqrt{ \frac{2 |\gS| |\gA|}{T} } - \varepsilon_{\rl}
    \\
    &= \max_{w \in \gW} \sum_{h=1}^H \sum_{(s, a) \in \gS \times \gA} w_h (s, a) \lp  \widetilde{d}^{\piE}_h(s, a) - d^{\widebar{\pi}, \widehat{P}}_h (s, a)  \rp - 2H \sqrt{ \frac{2 |\gS| |\gA|}{T} } - \varepsilon_{\rl}
    \\
    &= \lnorm \widetilde{d}^{\piE}_h - d^{\widebar{\pi}, \widehat{P}}_h \rnorm_{1} - 2H \sqrt{ \frac{2 |\gS| |\gA|}{T}} - \varepsilon_{\rl},
\end{align*}
where the last step again utilizes the dual representation of $\ell_1$-norm.  If we take $\varepsilon_{\rl} \leq \varepsilon/2$, $T \gtrsim H^2 |\gS||\gA|/\varepsilon^2$ and $\eta^{(t)} := \sqrt{|\gS||\gA| / (8T)}$, then we have
\begin{align*}
\sum_{h=1}^H \lnorm d^{\widebar{\pi}, \widehat{P}}_h - \widetilde{d}^{\piE}_h \rnorm_{1} \leq \min_{\pi \in \Pi} \sum_{h=1}^H \lnorm d^{\pi, \widehat{P}}_h - \widetilde{d}^{\piE}_h \rnorm_{1} + \varepsilon. 
\end{align*} 
We complete the proof.
\end{proof}

\subsection{Proof of Theorem \ref{theorem:sample-complexity-unknown-transition}}

\begin{proof}[Proof of Theorem \ref{theorem:sample-complexity-unknown-transition}]

Firstly, we verify assumption (a) in Proposition \ref{prop:connection}. With \cref{lem:reward_free}, when the number of trajectories collected by \textnormal{RF-Express} satisfies
\begin{align*}
    n \gtrsim  \frac{H^{3} |\gS| |\gA| }{\varepsilon^2}    \lp |\gS| + \log\lp\frac{|\gS| H}{\delta} \rp \rp,
\end{align*}
for any policy $\pi \in \Pi$ and reward function $w : \gS \times \gA \rar [0, 1]$, with probability at least $1-\delta/2$, $| V^{\pi, P, w} - V^{\pi, \widehat{P}, w} | \leq \varepsilon / 16 = \varepsilon_{\rfe}$. In a word, the assumption (a) in Proposition \ref{prop:connection} holds with $\delta_{\mathrm{RFE}} = \delta / 2$ and $\varepsilon_{\mathrm{RFE}} = \varepsilon / 16$.

Secondly, we note that the assumption (b) in Proposition \ref{prop:connection} holds by Lemma \ref{lemma:sample_complexity_of_new_estimator_unknown_transition}. More concretely, if the expert sample complexity and interaction complexity satisfies
\begin{align*}
    m \gtrsim   \frac{H^{3/2} | \gS | }{\varepsilon} \log\lp  \frac{|\gS| H}{\delta} \rp, \; n^\prime \gtrsim \frac{H^{2} | \gS |}{\varepsilon^2} \log\lp  \frac{|\gS| H}{\delta} \rp,
\end{align*}
with probability at least $1-\delta/2$, $\sum_{h=1}^H \Vert \widetilde{d}^{\piE}_h - d^{\piE}_h  \Vert_{1} \leq \varepsilon / 16 = \varepsilon_{\est}$. Hence, the assumption (b) in Proposition \ref{prop:connection} holds with $\delta_{\mathrm{EST}} = \delta / 2$ and $\varepsilon_{\mathrm{EST}} = \varepsilon / 16$.

Thirdly, we aim to verify that the assumption (c) in Proposition \ref{prop:connection} holds with $\widetilde{d}^{\piE}_h (s, a)$ and $\widehat{P}$. When $\varepsilon_{\rl} \leq \varepsilon / 2$ and $T \gtrsim |\gS| |\gA| H^2 / \varepsilon^2$ such that $2 H \sqrt{2 |\gS| |\gA| / T} \leq \varepsilon / 4$, we have that
\begin{align*}
    \sum_{h=1}^H \lnorm d^{\widebar{\pi}, \widehat{P}}_h - \widetilde{d}^{\piE}_h \rnorm_{1} - \min_{\pi \in \Pi} \sum_{h=1}^H \lnorm d^{\pi, \widehat{P}}_h - \widetilde{d}^{\piE}_h \rnorm_{1} \leq \frac{3\varepsilon}{4}  = \varepsilon_{\opt}.
\end{align*}
Therefore, the assumption (c) in Proposition \ref{prop:connection} holds with $\varepsilon_{\opt} = 3\varepsilon / 4$. Now, we summarize the conditions what we have obtained.
\begin{itemize}
    \item The assumption (a) in Proposition \ref{prop:connection} holds with $\delta_{\mathrm{RFE}} = \delta / 2$ and $\varepsilon_{\mathrm{RFE}} = \varepsilon / 16$.
    \item The assumption (b) in Proposition \ref{prop:connection} holds with $\delta_{\mathrm{EST}} = \delta / 2$ and $\varepsilon_{\mathrm{EST}} = \varepsilon / 16$.
    \item The assumption (c) in Proposition \ref{prop:connection} holds with $\varepsilon_{\opt} = 3\varepsilon / 4$. 
\end{itemize}
Applying Proposition \ref{prop:connection} finishes the proof. With probability at least $1-\delta$,
\begin{align*}
    V^{\piE} - V^{\widebar{\pi}} \leq 2 \varepsilon_{\mathrm{RFE}} + 2 \varepsilon_{\mathrm{EST}} + \varepsilon_{\opt} = \varepsilon.  
\end{align*}

\end{proof}

\section{Proof of Results in Section \ref{sec:mbtail_state_abstraction}}
\label{sec:proof_of_mbtail_state_abstraction}

\subsection{Reward-free Exploration Method with State Abstraction}
\label{subsec:appendix_rfe_method_state_abstraction}

\begin{algorithm}[h]
\caption{RF-Express with State Abstraction}
\label{algo:rf_express_state_abstraction}
\begin{algorithmic}[1]
\REQUIRE{A set of state abstractions $\{ \phi_h \}_{h=1}^H$, failure probability $\delta$, and function $\beta(n, \delta) = \log (3 \vert \Phi \vert \vert \gA \vert H / \delta)+ \vert \Phi \vert \log (8 e(n+1))$.}
\FOR{$t = 0, 1, 2, \cdots$}
\STATE{Update the abstract counter and abstract empirical transition model:}
\begin{align*}
    & n^{t}_h (x, a) = \sum_{i=1}^{t} \indict \{ \phi_h (s^{i}_h) = x, a^i_h = a \}, \; n^{t}_h (x, a, x^\prime) = \sum_{i=1}^{t} \indict \{ \phi_h (s^{i}_h) = x, a^i_h = a, \phi_{h+1} (s^{i}_{h+1}) = x^\prime  \},
    \\
    & \widehat{P}^{\phi, t}_h (x^\prime|x, a) = \frac{n^{t}_h (x, a, x^\prime)}{n^{t}_h (x, a)}, \; \text{if $n^{t}_h (x, a) > 0$ and } \widehat{P}^{\phi, t}_h (x^\prime|x, a) = \frac{1}{\vert \gS \vert}, \; \forall x^\prime \in \Phi \text{ otherwise}. 
\end{align*}
\STATE{Define $W^{t}_{H+1} (x, a) = 0, \; \forall (x, a) \in \Phi \times \gA$.}
\FOR{$h = H, H-1, \cdots, 1$}
\STATE{$W^{t}_h (x, a) = \min \left(H, 15 H^{2} \frac{\beta\left(n_{h}^{t } (x, a), \delta\right)}{n_{h}^{t } (x, a)}+\left(1+\frac{1}{H}\right) \sum_{x^{\prime} \in \Phi} \widehat{P}_{h}^{\phi, t} (x^{\prime} | x, a ) \max_{a^{\prime}} W_{h+1}^{t }\left(x^{\prime}, a^{\prime}\right)\right)$.}
\ENDFOR
\STATE{Derive the greedy policy: $\pi_{h}^{\phi, t+1}(x)=\argmax_{a \in \mathcal{A}} W_{h}^{ t}(x, a), \forall x \in \Phi, \forall h \in[H]$.}
\IF{$3 e \sqrt{ W_{1}^{t} ( \phi_1(s_1), \pi_{1}^{\phi, t+1} (\phi_1 (s_1))) }+ W_{1}^{t} (\phi_1(s_1), \pi_{1}^{\phi, t+1}( \phi_1 (s_1))) \leq \varepsilon / 2$}
\BREAK
\ENDIF
\STATE{Rollout $[\pi^{\phi, t+1}]^{M}$ to collect a trajectory $\tau^{t+1} = (s^{t+1}_1, a^{t+1}_1, s^{t+1}_2, a^{t+1}_2, \cdots, s^{t+1}_H, a^{t+1}_H)$. \label{alg_line:data_collection}}
\ENDFOR
\ENSURE{Transition model $\widehat{P}^{\phi, t}$.}
\end{algorithmic}
\end{algorithm}

\subsection{Problem Setup} 

To facilitate later analysis, we introduce some useful notations widely used in the literature \citep{li2006towards, jiang2015abstraction}. In this part, for a function $f$ that operates on the original state space $\gS$, we add a superscript $\phi$ (i.e., $f^{\phi}$) to denote the counterpart that operates on the abstract state space $\Phi$. Inversely, for a function $f^{\phi}$ that operates on the abstract state space, we use $[f^{\phi}]^{M}$ to denote its lifted version, which is defined as $[f^{\phi}]^{M} (s) = f^{\phi} (\phi (s))$. Notice that $[f^{\phi}]^{M}$ is a function over $\gS$.

\begin{defn}[Abstract MDP]
\label{def:abstract_mdp}
Under \cref{asmp:state_abstraction}, for the original MDP $\gM = (\gS, \gA, P, r, H, \rho)$, we define the abstract MDP $\gM^{\phi} = (\Phi, \gA, P^{\phi}, r^{\phi}, H, \rho^{\phi})$. In particular, 
\begin{itemize}
    \item $P^{\phi}_h (x^\prime | x, a) = \sum_{s^\prime \in \phi^{-1}_h (x^\prime)} P_h (s^\prime|s, a)$, for an arbitrary $s \in \phi^{-1}_h (x)$.
    \item $r^{\phi}_h (x, a) = r_h (s, a)$, for an arbitrary $s \in \phi^{-1}_h (x)$.
    \item $\rho^{\phi} (x) = \sum_{s \in \phi^{-1}_1 (x)} \rho (s, a)$.
\end{itemize}
Here $\phi^{-1}_h (x) = \{ s \in \gS: \phi_h (s) = x \}$. 
\end{defn}
We clarify that there is no ambiguity in \cref{def:abstract_mdp} because of \cref{asmp:state_abstraction}. The bisimulation condition enables that $s \in \phi_h^{-1}(x)$ are equivalent under the reward-consistent and transition-consistent conditions. With the abstract MDP $\gM^{\phi}$, for any abstract policy $\pi^{\phi}$, we utilize $V^{\pi^{\phi}, \gM^{\phi}}_h (x)$ to denote the corresponding value function. Similarly, with the original MDP $\gM$, for any policy $\pi$, we use $V^{\pi, \gM}_h (s)$ to denote the corresponding value function.

\begin{defn}[Abstract Expert Policy]
\label{def:abstract_expert_policy}
    Under \cref{asmp:state_abstraction}, for the original expert policy $\piE$, we define the abstract expert policy $\pi^{\expert, \phi}$. In particular, for any $(x, h) \in \Phi \times [H]$, it holds that
    \begin{align*}
        \pi^{\expert, \phi}_h (x) = \piE_h (s), \; \text{for an arbitrary } s \in \phi^{-1}_h (x). 
    \end{align*}
\end{defn}

Besides, for any policy $\pi \in \Pi$, we utilize $d^{\pi, \phi}_h \in \Delta (\Phi \times \gA)$ to denote the abstract state-action distribution.
\begin{align*}
    d^{\pi, \phi}_h (x, a) = \sP^{\pi} ( \phi_h (s_h) = x, a_h = a | P) = \sum_{s \in \phi^{-1}_h (x)} d^{\pi}_h (s, a). 
\end{align*}
For any abstract policy $\pi^{\phi} \in \Pi^{\phi}$ and abstract transition function $P^{\phi}$, we utilize $d^{\pi^{\phi}, P^{\phi}}_h \in \Delta (\Phi \times \gA)$ to denote the abstract state-action distribution induced by $\pi^{\phi}$ in $P^{\phi}$. In particular,
\begin{align*}
    d^{\pi^{\phi}, P^{\phi}}_h (x, a) = \sP^{\pi^{\phi}} (x_h = x, a_h = a| P^{\phi}).
\end{align*}

\subsection{Proof of Theorem \ref{theorem:sample-complexity-unknown-transition-state-abstraction}}

Prior to proving \cref{theorem:sample-complexity-unknown-transition-state-abstraction}, we provide a theoretical guarantee for the meta-algorithm presented in \cref{algo:framework_state_abstraction}. The algorithm constructs an abstract transition model, an abstract state-action distribution and an abstract policy. Finally, the algorithm outputs a policy that can operate in the original state space. To accomplish this, we introduce specialized analysis tools to connect these concepts in both the original and abstract spaces.      

\begin{prop}   \label{prop:connection_state_abstraction}
Suppose that 
\begin{itemize}  %
    \item[(a)] an algorithm A solves the reward-free exploration problem on the \emph{abstract} MDP $\gM^{\phi}$ (see Definition \ref{defn:reward_free}) up to an error $\varepsilon_{\rfe}$ with probability at least $1-\delta_{\rfe}$.
    \item[(b)] an algorithm B has an \emph{abstract} state-action distribution estimator for $d^{\piE, \phi}_h$, which satisfies $\sum_{h=1}^H \Vert \widetilde{d}^{\piE, \phi}_h - d^{\piE, \phi}_h  \Vert_{1} \leq \varepsilon_{\est}$, with probability at least $1-\delta_{\est}$;
    \item[(c)] with the \emph{abstract} transition model in (a) and the \emph{abstract} estimator in (b), an algorithm C solves the following optimization problem up to an error $\varepsilon_{\opt}$.
    \begin{align}
    \label{eq:ail_with_model_state_abstraction}
        \min_{\pi^\phi \in \Pi^{\phi}} \sum_{h=1}^H \lnorm \widetilde{d}^{\piE, \phi}_h - d^{\pi^{\phi}, \widehat{P}^{\phi}}_h \rnorm_1,
    \end{align}
    where $\Pi^{\phi}$ is the set of all abstract policies and $d^{\pi^{\phi}, \widehat{P}^{\phi}}_h$ is the abstract state-action distribution induced by the abstract policy $\pi^{\phi}$ and abstract transition function $\widehat{P}^{\phi}$.
\end{itemize}
Then applying algorithms A, B and C under the framework in Algorithm \ref{algo:framework_state_abstraction} could return a policy $[\widebar{\pi}^{\phi}]^{M}$, which has a policy value gap (i.e., $V^{\piE} - V^{[\widebar{\pi}^{\phi}]^{M}}$) at most $2 \varepsilon_{\est} + 2 \varepsilon_{\rfe} + \varepsilon_{\opt}$, with probability at least $1-\delta_{\est} - \delta_{\rfe}$.
\end{prop}
\begin{proof}
The proof idea is similar to that in \cref{subsec:proof_of_proposition_connection}. Additionally, we leverage the analysis techniques in state abstraction. We want to upper bound the imitation gap $V^{\piE, \gM} - V^{[\widebar{\pi}^{\phi}]^{M}, \gM}$, where $V^{\pi, \gM}$ represents the policy value of $\pi$ on the original MDP $\gM$. We consider the following two events.
\begin{align*}
    & E_{\est} = \lb \sum_{h=1}^H \lnorm \widetilde{d}^{\piE, \phi}_h - d^{\piE, \phi}_h  \rnorm_1 \leq \varepsilon_{\est}  \rb
    \\
    & E_{\rfe} = \lb \forall \pi^{\phi} \in \Pi^{\phi}, \; r^{\phi} = (r^{\phi}_1, \ldots, r^{\phi}_H), \; r^{\phi}_h: \Phi \times \gA \rar [0, 1] , \; | V^{\pi^{\phi}, P^{\phi}, r^{\phi}} - V^{\pi^{\phi}, \widehat{P}^{\phi}, r^{\phi}} | \leq \varepsilon_{\rfe} \rb. 
\end{align*}
With condition $(a)$ and condition $(b)$, we obtain $\sP (E_{\est} \cap E_{\rfe}) \geq 1 - \delta_{\rfe} - \delta_{\est}$. The following analysis is established on the event $E_{\est} \cap E_{\rfe}$.

By \cref{lemma:policy_value_irrelevant}, we have $V^{[\widebar{\pi}^{\phi}]^{M}, \gM} = V^{\widebar{\pi}^{\phi}, \gM^{\phi}}$, where $\gM^{\phi}$ is the abstract MDP in \cref{def:abstract_mdp}. Then we can upper bound the term $V^{\piE, \gM} - V^{\widebar{\pi}^{\phi}, \gM^{\phi}}$. On the event $E_{\rfe}$, we further have
\begin{align*}
    V^{\piE, \gM} - V^{\widebar{\pi}^{\phi}, \gM^{\phi}} &\leq  V^{\piE, \gM} - V^{\widebar{\pi}^{\phi}, \widehat{P}^{\phi}, r^{\phi}} + \varepsilon_{\rfe}
    \\
    &= \sum_{h=1}^H \sum_{(s, a) \in \gS \times \gA} d^{\piE}_h (s, a) r_h (s, a) - \sum_{h=1}^H \sum_{(x, a) \in \Phi \times \gA} d^{\widebar{\pi}^{\phi}, \widehat{P}^{\phi}}_h (x, a) r^{\phi}_h (x, a)+ \varepsilon_{\rfe}
    \\
    &= \sum_{h=1}^H \sum_{(x, a) \in \Phi \times \gA} d^{\piE, \phi}_h (x, a) r^{\phi}_h (x, a) - \sum_{h=1}^H \sum_{(x, a) \in \Phi \times \gA} d^{\widebar{\pi}^{\phi}, \widehat{P}^{\phi}}_h (x, a) r^{\phi}_h (x, a)+ \varepsilon_{\rfe} . 
\end{align*}
Here $d^{\widebar{\pi}^{\phi}, \widehat{P}^{\phi}}_h$ is the abstract state-action distribution of $\widebar{\pi}^{\phi}$ in $\widehat{P}^{\phi}$ and $d^{\piE, \phi}_h (x, a) = \sum_{s \in \phi^{-1}_h (x)} d^{\piE}_h (s, a)$. The last equation holds due to the reward-consistent condition in \eqref{eq:bisimulation}. Then we can obtain
\begin{align*}
    V^{\piE, \gM} - V^{\widebar{\pi}^{\phi}, \gM^{\phi}} &\leq  \sum_{h=1}^H \sum_{(x, a) \in \Phi \times \gA} d^{\piE, \phi}_h (x, a) r^{\phi}_h (x, a) - \sum_{h=1}^H \sum_{(x, a) \in \Phi \times \gA} d^{\widebar{\pi}^{\phi}, \widehat{P}^{\phi}}_h (x, a) r^{\phi}_h (x, a)+ \varepsilon_{\rfe}
    \\
    &\overset{(a)}{\leq} \sum_{h=1}^H \lnorm d^{\piE, \phi}_h - d^{\widebar{\pi}^{\phi}, \widehat{P}^{\phi}}_h \rnorm_1+ \varepsilon_{\rfe}
    \\
    &\leq \sum_{h=1}^H \lnorm d^{\piE, \phi}_h  - \widetilde{d}^{\piE, \phi}_h  \rnorm_1 + \sum_{h=1}^H \lnorm \widetilde{d}^{\piE, \phi}_h  - d^{\widebar{\pi}^{\phi}, \widehat{P}^{\phi}}_h \rnorm_1+ \varepsilon_{\rfe}
    \\
    &\overset{(b)}{\leq}  \sum_{h=1}^H \lnorm \widetilde{d}^{\piE, \phi}_h - d^{\widebar{\pi}^{\phi}, \widehat{P}^{\phi}}_h  \rnorm_1 + \varepsilon_{\est}+ \varepsilon_{\rfe}. 
\end{align*}
Inequality $(a)$ holds due to the dual representation of $\ell_1$-norm and inequality $(b)$ holds due to the event $E_{\est}$. Because $\widebar{\pi}^{\phi}$ is an $\varepsilon_{\opt}$-optimal solution of the optimization problem in \eqref{eq:ail_with_model_state_abstraction}, we get that
\begin{align*}
    V^{\piE, \gM} - V^{\widebar{\pi}^{\phi}, \gM^{\phi}} &\leq \min_{\pi^{\phi} \in \Pi^{\phi}} \sum_{h=1}^H \lnorm \widetilde{d}^{\piE, \phi}_h  - d^{\pi^{\phi}, \widehat{P}^{\phi}}_h  \rnorm_1 + \varepsilon_{\est}+ \varepsilon_{\rfe} + \varepsilon_{\opt}. 
\end{align*}
We consider the abstract expert policy $\pi^{\expert, \phi}$ in \cref{def:abstract_expert_policy}. Since $\pi^{\expert, \phi} \in \Pi^{\phi}$, it holds that
\begin{align*}
    V^{\piE, \gM} - V^{\widebar{\pi}^{\phi}, \gM^{\phi}} &\leq \sum_{h=1}^H \lnorm \widetilde{d}^{\piE, \phi}_h  - d^{\pi^{\expert, \phi}, \widehat{P}^{\phi}}_h  \rnorm_1 + \varepsilon_{\est}+ \varepsilon_{\rfe} + \varepsilon_{\opt}
    \\
    &\leq \sum_{h=1}^H \lnorm \widetilde{d}^{\piE, \phi}_h - d^{\pi^{\expert, \phi}, P^{\phi}}_h  \rnorm_1 + \sum_{h=1}^H \lnorm d^{\pi^{\expert, \phi}, P^{\phi}}_h  - d^{\pi^{\expert, \phi}, \widehat{P}^{\phi}}_h  \rnorm_1 + \varepsilon_{\est}+ \varepsilon_{\rfe} + \varepsilon_{\opt} 
\end{align*}
Then we upper bound the term $\sum_{h=1}^H \Vert d^{\pi^{\expert, \phi}, P^{\phi}}_h - d^{\pi^{\expert, \phi}, \widehat{P}^{\phi}}_h  \Vert_1$
\begin{align*}
    \sum_{h=1}^H \lnorm d^{\pi^{\expert, \phi}, P^{\phi}}_h - d^{\pi^{\expert, \phi}, \widehat{P}^{\phi}}_h  \rnorm_1 &= \max_{r^{\phi} \in \gW^{\phi}} \sum_{h=1}^H \sum_{(x, a) \in \Phi \times \gA} \lp d^{\pi^{\expert, \phi}, P^{\phi}}_h (x, a) - d^{\pi^{\expert, \phi}, \widehat{P}^{\phi}}_h (x, a) \rp r^{\phi}_h (x, a)
    \\
    &= \max_{r^{\phi} \in \gW^{\phi}} V^{\pi^{\expert, \phi}, P^{\phi}, r^{\phi}} - V^{\pi^{\expert, \phi}, \widehat{P}^{\phi}, r^{\phi}}
    \\
    &\leq \varepsilon_{\rfe}.
\end{align*}
Here $\gW^{\phi} = \{ w^{\phi} = (w^{\phi}_1, \ldots, w^{\phi}_H), \; w^{\phi}_h: \Phi \times \gA \rar [0, 1], \forall h \in [H] \}$. The last inequality holds due to the event $E_{\rfe}$. Then we obtain
\begin{align*}
    V^{\piE, \gM} - V^{\widebar{\pi}^{\phi}, \gM^{\phi}} &\leq \sum_{h=1}^H \lnorm \widetilde{d}^{\piE, \phi}_h - d^{\pi^{\expert, \phi}, P^{\phi}}_h  \rnorm_1  + \varepsilon_{\est}+ 2\varepsilon_{\rfe} + \varepsilon_{\opt}. 
\end{align*}
Applying \cref{lemma:state_action_distribution_irrelevant} on $\pi^{\expert, \phi}$ and $P^{\phi}$ yields $d^{\pi^{\expert, \phi}, P^{\phi}}_h = d^{[\pi^{\expert, \phi}]^M, P, \phi}_h$. Combined with $[\pi^{\expert, \phi}]^M = \piE$ in \cref{lemma:reward_transition_irrelevant}, we obtain
\begin{align*}
    V^{\piE, \gM} - V^{\widebar{\pi}^{\phi}, \gM^{\phi}} &\leq \sum_{h=1}^H \lnorm \widetilde{d}^{\piE, \phi}_h  - d^{\pi^{\expert, \phi}, P^{\phi}}_h  \rnorm_1  + \varepsilon_{\est}+ 2\varepsilon_{\rfe} + \varepsilon_{\opt}
    \\
    &= \sum_{h=1}^H \lnorm \widetilde{d}^{\piE, \phi}_h - d^{\piE, \phi}_h \rnorm_1  + \varepsilon_{\est}+ 2\varepsilon_{\rfe} + \varepsilon_{\opt}
    \\
    &\leq 2\varepsilon_{\est}+ 2\varepsilon_{\rfe} + \varepsilon_{\opt}, 
\end{align*}
where the last inequality holds due to the event $E_{\est}$. We finish the proof.
\end{proof}

Now, we proceed to prove Theorem \ref{theorem:sample-complexity-unknown-transition-state-abstraction}.  
\begin{proof}[Proof of Theorem \ref{theorem:sample-complexity-unknown-transition-state-abstraction}]
First, we verify condition $(a)$ in Proposition \ref{prop:connection_state_abstraction}. We want to demonstrate that Algorithm \ref{algo:rf_express_state_abstraction} is equivalent to applying RF-Express (Algorithm \ref{algo:rf_express}) on the abstract MDP $\gM^{\phi}$. The only difference lies in the data-collection process. On one hand, in line \ref{alg_line:data_collection} in Algorithm \ref{algo:rf_express_state_abstraction}, we roll out the lifted policy $[\pi^{\phi, t+1}]^{M}$ on the original MDP $\gM$. On the other hand, when applying RF-Express (Algorithm \ref{algo:rf_express}) on the abstract MDP $\gM^{\phi}$, we rollout the abstract policy $\pi^{\phi, t+1}$ on the abstract MDP $\gM^{\phi}$. We will prove that in the above two data-collection processes, the corresponding abstract-state-action distributions are actually the same. Consequently, Algorithm \ref{algo:rf_express_state_abstraction} can be regarded as applying RF-Express (Algorithm \ref{algo:rf_express}) on the abstract MDP $\gM^{\phi}$.

In the first process, conditioned on $\pi^{\phi, t+1}$, we consider the probability distribution of $(\phi_h (s^{t+1}_h), a^{t+1}_h)$. Recall the definition: 
\begin{align*}
    d^{[\pi^{\phi, t+1}]^{M}, P, \phi}_h (x, a) := 
    \sP \lp \phi_h (s^{t+1}_h) = x, a^{t+1}_h = a | [\pi^{\phi, t+1}]^{M}, P \rp = \sum_{s \in \phi^{-1}_h (x)} \sP \lp s^{t+1}_h = s, a^{t+1}_h = a | [\pi^{\phi, t+1}]^{M}, P \rp. 
\end{align*}
By \cref{lemma:state_action_distribution_irrelevant}, we have that
\begin{align*}
    d^{[\pi^{\phi, t+1}]^{M}, P, \phi}_h (x, a) = d^{\pi^{\phi, t+1}, P^{\phi}}_h (x, a).
\end{align*}
Notice that the distribution $d^{\pi^{\phi, t+1}, P^{\phi}}_h (x, a)$ is exactly the abstract state-action distribution of $\pi^{\phi, t+1}$ in the abstract MDP $\gM^{\phi}$. Therefore, in the mentioned two data-collection processes, the corresponding abstract-state-action distributions are actually the same. Then we can apply \cref{lem:reward_free} on the abstract MDP. When the number of trajectories collected by Algorithm \ref{algo:rf_express_state_abstraction} satisfies
\begin{align*}
    n \gtrsim  \frac{H^{3} |\Phi| |\gA| }{\varepsilon^2}    \lp |\Phi| + \log\lp\frac{|\Phi| H}{\delta} \rp \rp,
\end{align*}
for any policy $\pi^{\phi} \in \Pi^{\phi}$ and reward function $r^{\phi} = (r^{\phi}_1, \ldots, r^{\phi}_H), \; r^{\phi}_h: \Phi \times \gA \rar [0, 1]$, with probability at least $1-\delta/2$, $| V^{\pi^{\phi}, P^{\phi}, r^{\phi}} - V^{\pi^{\phi}, \widehat{P}^{\phi}, r^{\phi}} | \leq \varepsilon / 16 = \varepsilon_{\rfe}$. In summary, the assumption (a) in Proposition \ref{prop:connection_state_abstraction} holds with $\delta_{\mathrm{RFE}} = \delta / 2$ and $\varepsilon_{\mathrm{RFE}} = \varepsilon / 16$.

Second, we verify the condition $(b)$ in Proposition \ref{prop:connection_state_abstraction}. Note that the assumption (b) in Proposition \ref{prop:connection_state_abstraction} holds by Lemma \ref{lemma:sample_complexity_of_new_estimator_unknown_transition_state_abstraction}. More concretely, if the expert sample complexity and interaction complexity satisfies
\begin{align*}
    m \gtrsim   \frac{H^{3/2} | \Phi | }{\varepsilon} \log\lp  \frac{|\Phi| H}{\delta} \rp, \; n^\prime \gtrsim \frac{H^{2} | \Phi |}{\varepsilon^2} \log\lp  \frac{|\Phi| H}{\delta} \rp,
\end{align*}
with probability at least $1-\delta/2$, $\sum_{h=1}^H \Vert \widetilde{d}^{\piE, \phi}_h - d^{\piE, \phi}_h  \Vert_{1} \leq \varepsilon / 16 = \varepsilon_{\est}$. Hence, the assumption (b) in Proposition \ref{prop:connection_state_abstraction} holds with $\delta_{\mathrm{EST}} = \delta / 2$ and $\varepsilon_{\mathrm{EST}} = \varepsilon / 16$.

Third, we validate the condition $(c)$ in Proposition \ref{prop:connection_state_abstraction}. In particular, we apply \cref{algo:gradient_based_optimization} to solve the following abstract state-action distribution matching problem.
\begin{align*}
        \min_{\pi^\phi \in \Pi^{\phi}} \sum_{h=1}^H \lnorm \widetilde{d}^{\piE, \phi}_h - d^{\pi^{\phi}, \widehat{P}^{\phi}}_h \rnorm_1.
\end{align*}
Therefore, we can apply \cref{lemma:approximate-minimax}. In particular, when $\varepsilon_{\rl} \leq \varepsilon / 2$ and $T \gtrsim |\Phi| |\gA| H^2 / \varepsilon^2$ such that $2 H \sqrt{2 |\Phi| |\gA| / T} \leq \varepsilon / 4$, we have that
\begin{align*}
    \sum_{h=1}^H \lnorm \widetilde{d}^{\piE, \phi}_h - d^{\widebar{\pi}^{\phi}, \widehat{P}^{\phi}}_h \rnorm_1 - \min_{\pi^\phi \in \Pi^{\phi}} \sum_{h=1}^H \lnorm \widetilde{d}^{\piE, \phi}_h - d^{\pi^{\phi}, \widehat{P}^{\phi}}_h \rnorm_1 \leq \frac{3\varepsilon}{4}  = \varepsilon_{\opt}.
\end{align*}
In summary, we have established the following conditions:
\begin{itemize}
    \item Assumption (a) in Proposition \ref{prop:connection_state_abstraction} holds with $\delta_{\mathrm{RFE}} = \delta / 2$ and $\varepsilon_{\mathrm{RFE}} = \varepsilon / 16$.
    \item Assumption (b) in Proposition \ref{prop:connection_state_abstraction} holds with $\delta_{\mathrm{EST}} = \delta / 2$ and $\varepsilon_{\mathrm{EST}} = \varepsilon / 16$.
    \item Assumption (c) in Proposition \ref{prop:connection_state_abstraction} holds with $\varepsilon_{\opt} = 3\varepsilon / 4$. 
\end{itemize}
By applying Proposition \ref{prop:connection_state_abstraction}, we complete the proof. With probability at least $1-\delta$, we have
\begin{align*}
    V^{\piE} - V^{[\widebar{\pi}^{\phi}]^M} \leq 2 \varepsilon_{\mathrm{RFE}} + 2 \varepsilon_{\mathrm{EST}} + \varepsilon_{\opt} = \varepsilon.  
\end{align*}
\end{proof}

\subsection{Useful Lemmas}
In this part, we develop specialized analysis tools for AIL with state abstraction. The below lemma indicates that under \cref{asmp:state_abstraction}, the lifted versions of the abstract reward function and abstract transition function are identical to the original reward function and transition function, respectively.   
\begin{lem}
\label{lemma:reward_transition_irrelevant}
For the original MDP $\gM = (\gS, \gA, P, r, H, \rho)$ and expert policy $\piE$ that satisfy \cref{asmp:state_abstraction}, we consider the abstract MDP $\gM^{\phi} = (\Phi, \gA, P^{\phi}, r^{\phi}, H, \rho^{\phi})$ in \cref{def:abstract_mdp}. Then we have that
\begin{align*}
\forall h \in [H], \; (s, a) \in \gS \times \gA, \; x^\prime \in \Phi, \; r_h (s, a) = [r^{\phi}]^{M}_h (s, a), \; \sum_{s^\prime \in \phi^{-1}_{h+1} (x^\prime)} P_h (s^\prime |s, a) = [P^{\phi}]^{M}_h (x^\prime|s, a).    
\end{align*}
Here $[r^{\phi}]^{M}_h (s, a) = r^{\phi}_h (\phi_h (s), a)$ and $[P^{\phi}]^{M}_h (x^\prime|s, a) = P^{\phi}_h (x^\prime | \phi_h (s), a)$. Furthermore, we consider the abstract expert policy $\pi^{\expert, \phi}$ in \cref{def:abstract_expert_policy}. Then we have that 
\begin{align*}
    \forall h \in [H], \; s \in \gS, \; \piE_h (s) = [\pi^{\expert, \phi}]^{M}_h (s), 
\end{align*}
where $[\pi^{\expert, \phi}]^{M}_h (s) = \pi^{\expert, \phi}_h (\phi_h (s)) $.
\end{lem}

\begin{proof}
For the reward function, we have
\begin{align*}
    [r^{\phi}_h]_{\gM} (s, a) = r^{\phi}_h (\phi_h (s), a) \overset{x:= \phi_h (s)}{=} r^{\phi}_h (x, a). 
\end{align*}
Notice that $r^{\phi}_h (x, a) = r_h (\widehat{s}, a)$ for an arbitrary $\widehat{s} \in \phi^{-1}_h (x)$. Moreover, since $s, \; \widehat{s} \in \phi^{-1}_h (x)$ and $r$ satisfies \eqref{eq:bisimulation}, we have $r_h (\widehat{s}, a) = r_h (s, a)$. 

For the transition function, we have
\begin{align*}
    [P^{\phi}]^{M}_h (x^\prime|s, a) = P^{\phi}_h (x^\prime | \phi_h (s), a) \overset{x := \phi_h (s)}{=} P^{\phi}_h (x^\prime | x, a).  
\end{align*}
According to \cref{def:abstract_mdp}, we have $P^{\phi}_h (x^\prime | x, a) = \sum_{s^\prime \in \phi^{-1}_{h+1} (x^\prime)} P_h (s^\prime| \widetilde{s}, a)$ for an arbitrary $\widetilde{s} \in \phi^{-1}_h (x)$. Furthermore, because $s, \; \widetilde{s} \in \phi^{-1}_h (x)$ and $P$ satisfies \eqref{eq:bisimulation}, we have 
\begin{align*}
\sum_{s^\prime \in \phi^{-1}_{h+1} (x^\prime)} P_h (s^\prime| \widetilde{s}, a) = \sum_{s^\prime \in \phi^{-1}_{h+1} (x^\prime)} P_h (s^\prime| s, a).
\end{align*}
Finally, for the expert policy, it holds that 
\begin{align*}
    [\pi^{\expert, \phi}]^{M}_h (s) = \pi^{\expert, \phi}_h (\phi_h (s)) \overset{x:= \phi_h (s)}{=} \pi^{\expert, \phi}_h (x).  
\end{align*}
According to \cref{def:abstract_expert_policy}, we have $\pi^{\expert, \phi}_h (x) = \piE_h (\widetilde{s})$ for an arbitrary $\widetilde{s} \in \phi^{-1}_h (x)$. Notice that $s, \; \widetilde{s} \in \phi^{-1}_h (x)$ and $\piE$ satisfies \eqref{eq:expert_consistent}. Therefore, we have $\pi^{\expert, \phi}_h (x) = \piE_h (s)$. We finish the proof.
\end{proof}

\begin{lem}
\label{lemma:state_abstraction_summation}
For any function $f: \Phi \rightarrow \reals$, $g: \gS \rightarrow \reals$ and an state abstraction $\phi: \gS \rightarrow \Phi$, we define $g^{\phi} (x) := \sum_{s \in \phi^{-1}(x)} g (s)$, then we have
\begin{align*}
    \sum_{x \in \Phi} g^{\phi} (x) f(x) = \sum_{s \in \gS } g(s) [f]^{M} (s), 
\end{align*}
where $[f]^{M} (s) = f (\phi (s))$.
\end{lem}
\begin{proof}
    \begin{align*}
    \sum_{x \in \Phi} g^{\phi} (x) f(x) &= \sum_{x \in \Phi} \sum_{s \in \phi^{-1}(x)} g (s) f(x)
    \\
    &= \sum_{x \in \Phi} \sum_{s \in \gS} \indict \lb s \in \phi^{-1}(x)  \rb g (s) f(x)
    \\
    &= \sum_{s \in \gS} \sum_{x \in \Phi}  \indict \lb x = \phi (s)    \rb g (s) f(x)
    \\
    &= \sum_{s \in \gS} g (s) f(\phi (s))
    \\
    &= \sum_{s \in \gS} g (s) [f]^{M} (s). 
\end{align*}
We complete the proof.
\end{proof}

\cref{lemma:policy_value_irrelevant} indicates that for any abstract policy $\pi^{\phi} \in \Pi^{\phi}$, the value function of $[\pi^{\phi}]^M$ on $P$ equals the lifted version of the value function of $\pi^{\phi}$ on $P^{\phi}$.
\begin{lem}
\label{lemma:policy_value_irrelevant}
For the original MDP $\gM = (\gS, \gA, P, r, H, \rho)$ and expert policy $\piE$ that satisfy \cref{asmp:state_abstraction}, we consider the abstract MDP $\gM^{\phi} = (\Phi, \gA, P^{\phi}, r^{\phi}, H, \rho^{\phi})$ in \cref{def:abstract_mdp}. Then, for any abstract policy $\pi^{\phi} \in \Pi^{\phi}$, we have
\begin{align*}
V^{[\pi^{\phi}]^{M}, \gM}_h (s) = [V^{\pi^{\phi}, \gM^{\phi}}]^{M}_h (s), \forall s \in \gS, h \in [H],    
\end{align*}
where $[V^{\pi^{\phi}, \gM^{\phi}}]^{M}_h (s) := V^{\pi^{\phi}, \gM^{\phi}}_h (\phi_h (s))$, $[\pi^{\phi}]^{M}_h (a|s) = \pi^{\phi}_h (a|\phi_h (s))$. $V^{\pi^{\phi}, \gM^{\phi}}_h (s)$ is the value function of $\pi^{\phi}$ on $\gM^{\phi}$ and $V^{[\pi^{\phi}]^{M}, \gM}_h (s)$ is the value function of $[\pi^{\phi}]^{M}$ on $\gM$. Furthermore, it holds that $V^{[\pi^{\phi}]^{M}, \gM} = V^{\pi^{\phi}, \gM^{\phi}}$.
\end{lem}

\begin{proof}
The proof is based on backward induction. For the base case, we prove that
\begin{align*}
    V^{[\pi^{\phi}]^{M}, \gM}_H (s) = [V^{\pi^{\phi}, \gM^{\phi}}]^{M}_H (s), \; \forall s \in \gS.
\end{align*}
In particular,
\begin{align*}
    [V^{\pi^{\phi}, \gM^{\phi}}]^{M}_H (s) &= V^{\pi^{\phi}, \gM^{\phi}}_H (\phi_H (s))
    \\
    &= \sum_{a \in \gA} \pi^{\phi}_H (a| \phi_H (s)) r^{\phi}_H (\phi_H (s), a)
    \\
    &= \sum_{a \in \gA} [\pi^{\phi}]^{M}_H (a| s) [r^{\phi}]^{M}_H (s, a)
    \\
    &\overset{(a)}{=}  \sum_{a \in \gA} [\pi^{\phi}]^{M}_H (a| s) r_H (s, a)
    \\
    &= V^{[\pi^{\phi}]^{M}, \gM}_{H} (s).
\end{align*}
Equation $(a)$ follows \cref{lemma:reward_transition_irrelevant}. We finish the proof of the base case and continue to prove the induction stage. Assume that $V^{[\pi^{\phi}]^{M}, \gM}_{h+1} (s) = [V^{\pi^{\phi}, \gM^{\phi}}]^{M}_{h+1} (s), \forall s \in \gS$, we consider the time step $h$.

\begin{align*}
    [V^{\pi^{\phi}, \gM^{\phi}}]^{M}_{h} (s) &= V^{\pi^{\phi}, \gM^{\phi}}_{h} (\phi_h (s))
    \\
    &= \expect_{a \sim  \pi^{\phi}_h (\cdot|\phi_h (s))} \ls r^{\phi}_h (\phi_h (s), a) + P^{\phi}_{h+1} V^{\pi^{\phi}, \gM^{\phi}}_{h+1} (\phi_h (s), a) \rs.
\end{align*}
Here $P^{\phi}_{h+1} V^{\pi^{\phi}, \gM^{\phi}}_{h+1} (\phi_h (s), a) = \expect_{x^\prime \sim P^{\phi}_{h+1}(\cdot|\phi_h (s), a )} \ls V^{\pi^{\phi}, \gM^{\phi}}_{h+1} (x^\prime) \rs$. For the first term in RHS, we have

\begin{align*}
    \expect_{a \sim  \pi^{\phi}_h (\cdot|\phi_h (s))} \ls r^{\phi}_h (\phi_h (s), a) \rs = \expect_{a \sim  [\pi^{\phi}]^{M}_h (\cdot|s)} \ls [r^{\phi}]^{M}_h (s, a) \rs = \expect_{a \sim  [\pi^{\phi}]^{M}_h (\cdot|s)} \ls r_h (s, a) \rs. 
\end{align*}
The last equation utilizes \cref{lemma:reward_transition_irrelevant}. For the term $P^{\phi}_{h+1} V^{\pi^{\phi}, \gM^{\phi}}_{h+1} (\phi_h (s), a)$, we obtain

\begin{align*}
    P^{\phi}_{h+1} V^{\pi^{\phi}, \gM^{\phi}}_{h+1} (\phi_h (s), a) &= \sum_{x^\prime \in \Phi} P^{\phi}_{h+1} (x^\prime|\phi_h (s), a) V^{\pi^{\phi}, \gM^{\phi}}_{h+1} (x^\prime)
    \\
    &= \sum_{x^\prime \in \Phi} \lp \sum_{s^\prime \in \phi^{-1}_h (x^\prime)} P_{h+1} (s^\prime|\phi_h (s), a) \rp V^{\pi^{\phi}, \gM^{\phi}}_{h+1} (x^\prime), 
    \\
    &= \sum_{x^\prime \in \Phi} \lp \sum_{s^\prime \in \phi^{-1}_h (x^\prime)} P_{h+1} (s^\prime|\tilde{s}, a) \rp V^{\pi^{\phi}, \gM^{\phi}}_{h+1} (x^\prime), \; \text{for an arbitrary } \widetilde{s} \in \phi^{-1}_h (x). 
\end{align*}
In the last equation, we define $x = \phi_h (s)$. According to $s, \; \widetilde{s} \in \phi^{-1}_h (x)$ and \eqref{eq:bisimulation} in \cref{asmp:state_abstraction}, we have 

\begin{align*}
    P^{\phi}_{h+1} V^{\pi^{\phi}, \gM^{\phi}}_{h+1} (\phi_h (s), a) &= \sum_{x^\prime \in \Phi} \lp \sum_{s^\prime \in \phi^{-1}_h (x^\prime)} P_{h+1} (s^\prime|\tilde{s}, a) \rp V^{\pi^{\phi}, \gM^{\phi}}_{h+1} (x^\prime)
    \\
    &= \sum_{x^\prime \in \Phi}  \lp \sum_{s^\prime \in \phi^{-1}_h (x^\prime)} P_{h+1} (s^\prime|s, a) \rp V^{\pi^{\phi}, \gM^{\phi}}_{h+1} (x^\prime). 
\end{align*}
Applying \cref{lemma:state_abstraction_summation} with $f (x) = V^{\pi^{\phi}, \gM^{\phi}}_{h+1} (x), g (s^\prime) = P_{h+1} (s^\prime|s, a), \phi = \phi_{h+1}$ yields that
\begin{align*}
    P^{\phi}_{h+1} V^{\pi^{\phi}, \gM^{\phi}}_{h+1} (\phi_h (s), a) &= \sum_{s^\prime \in \gS} P_{h+1} (s^\prime|s, a) \ls V^{\pi^{\phi}, \gM^{\phi}} \rs^{M}_{h+1} (s^\prime)
    \\
    &\overset{(a)}{=} \sum_{s^\prime \in \gS} P_{h+1} (s^\prime|s, a) V^{[\pi^{\phi}]^{M}, \gM}_{h+1} (s^\prime)
    \\
    &= P_{h+1} V^{[\pi^{\phi}]^{M}, \gM}_{h+1} (s, a). 
\end{align*}
In equation $(a)$, we leverage the assumption in time step $h+1$. Then we obtain
\begin{align*}
    [V^{\pi^{\phi}, \gM^{\phi}}]^{M}_{h} (s) &= \expect_{a \sim  \pi^{\phi}_h (\cdot|\phi_h (s))} \ls r^{\phi}_h (\phi_h (s), a) + P^{\phi}_{h+1} V^{\pi^{\phi}, \gM^{\phi}}_{h+1} (\phi_h (s), a) \rs
    \\
    &= \expect_{a \sim  [\pi^{\phi}]^{M}_h (\cdot|s)} \ls r_h (s, a) + P_{h+1} V^{[\pi^{\phi}]^{M}, \gM}_{h+1} (s, a) \rs
    \\
    &= V^{[\pi^{\phi}]^{M}, \gM}_{h} (s). 
\end{align*}
We prove the induction stage and thus finish the proof of the first claim. Furthermore, according to the definition of $\rho^{\phi}$, we have

\begin{align*}
    V^{\pi^{\phi}, \gM^{\phi}} = \expect_{x \sim \rho^{\phi}} \ls  V^{\pi^{\phi}, \gM^{\phi}}_1 (x)  \rs = \sum_{x \in \Phi} \rho^{\phi} (x) V^{\pi^{\phi}, \gM^{\phi}}_1 (x) = \sum_{s \in \gS} \rho (s) \ls V^{\pi^{\phi}, \gM^{\phi}} \rs^{M}_1 (s).  
\end{align*}
In the last equation, we apply \cref{lemma:state_abstraction_summation} with $f (x) = V^{\pi^{\phi}, \gM^{\phi}}_1 (x)$, $g (s) = \rho (s)$ and $\phi = \phi_1$. We have proved that $[ V^{\pi^{\phi}, \gM^{\phi}} ]^{M}_1 (s) = V^{[\pi^{\phi}]^{M}, \gM}_{1} (s)$. Then it holds that
\begin{align*}
    V^{\pi^{\phi}, \gM^{\phi}} = \sum_{s \in \gS} \rho (s) V^{[\pi^{\phi}]^{M}, \gM}_{1} (s) = V^{[\pi^{\phi}]^{M}, \gM}_{1},  
\end{align*}
which completes the proof.
\end{proof}

\begin{lem}
\label{lemma:state_action_distribution_irrelevant}
For the original MDP $\gM = (\gS, \gA, P, r, H, \rho)$ and expert policy $\piE$ that satisfy \cref{asmp:state_abstraction}, we consider the abstract MDP $\gM^{\phi} = (\Phi, \gA, P^{\phi}, r^{\phi}, H, \rho^{\phi})$ in \cref{def:abstract_mdp}. Then, for any abstract policy $\pi^{\phi} \in \Pi^{\phi}$,
\begin{align*}
    \forall h \in [H], \; (x, a) \in \Phi \times 
    \gA, \; d^{\pi^{\phi}, P^{\phi}}_h (x, a) = d^{[\pi^{\phi}]^{M}, P, \phi}_h (x, a).
\end{align*}
Here $d^{\pi^{\phi}, P^{\phi}}_h (x, a) = \sP ( x_h = x, a_h = a | \pi^{\phi}, P^{\phi} )$ and $d^{[\pi^{\phi}]^{M}, P, \phi}_h (x, a) = \sP (\phi_h (s_h) = x, a_h = a | [\pi^{\phi}]^{M}, P) = \sum_{s \in \phi^{-1}_h (x)} d^{[\pi^{\phi}]^{M}, P}_h (s, a)$.
\end{lem}

\begin{proof}
We first prove that for any fixed $x \in \Phi, h \in [H]$, 
\begin{align*}
    d^{\pi^{\phi}, P^{\phi}}_h (x) = d^{[\pi^{\phi}]^{M}, P, \phi}_h (x),
\end{align*}
where $d^{\pi^{\phi}, P^{\phi}}_h (x) = \sP \lp x_h = x | \pi^{\phi}, P^{\phi}  \rp$ and $d^{[\pi^{\phi}]^{M}, P, \phi}_h (x) = \sP \lp \phi_h (s_h) = x | [\pi^{\phi}]^{M}, P  \rp$. Consider any fixed $x \in \Phi, h \in [H]$, we construct an abstract reward function $\widetilde{r}^{\phi}$.
\begin{align*}
    &\widetilde{r}^{\phi}_{h} (x, a) = 1, \forall a \in \gA,
    \\
    &\widetilde{r}^{\phi}_{\ell} (\tilde{x}, a) = 0, \forall \tilde{x} \in \Phi \setminus \{x \}, a \in \gA, \ell \in [H] \setminus \{ h \}. 
\end{align*}
Furthermore, we consider $[\widetilde{r}^{\phi}]^{M}$, which is the lifted version of $\widetilde{r}^{\phi}$.
\begin{align*}
    & \ls \widetilde{r}^{\phi} \rs^{M}_{h} (s, a) = 1, \forall s \in \phi^{-1}_h (x), a \in \gA,
    \\
    &\ls \widetilde{r}^{\phi} \rs^{M}_{\ell} (s, a) = 0, \forall s \in \gS \setminus \phi^{-1}_h (x), a \in \gA, \ell \in [H] \setminus \{ h \}. 
\end{align*}

On the one hand, according to the dual formulation of policy value in \eqref{eq:dual_of_policy_value}, we can get that $d^{\pi^{\phi}, P^{\phi}}_h (x) = V^{\pi^{\phi}, P^{\phi}, \widetilde{r}^{\phi}}$. On the other hand, it holds that

\begin{align*}
    d^{[\pi^{\phi}]^{M}, P, \phi}_h (x) = \sum_{s \in \phi^{-1}_h (x)} d^{[\pi^{\phi}]^{M}, P}_h (s) = V^{[\pi^{\phi}]^{M}, P, [\widetilde{r}^{\phi}]^{M}}.
\end{align*}

The last equation still follows the dual representation of policy value. Notice that $[\widetilde{r}^{\phi}]^{M}$ satisfies the reward-consistent condition (i.e., \eqref{eq:bisimulation} in \cref{asmp:state_abstraction}). With \cref{lemma:policy_value_irrelevant}, we get that $V^{[\pi^{\phi}]^{M}, P, [\widetilde{r}^{\phi}]^{M}} = V^{\pi^{\phi}, P^{\phi}, \widetilde{r}^{\phi}}$, which implies that $d^{\pi^{\phi}, P^{\phi}}_h (x) = d^{[\pi^{\phi}]^{M}, P, \phi}_h (x)$. Then we have that
\begin{align*}
    d^{\pi^{\phi}, P^{\phi}}_h (x, a) = d^{\pi^{\phi}, P^{\phi}}_h (x) \pi^{\phi}_h (a|x) = d^{[\pi^{\phi}]^{M}, P, \phi}_h (x) \pi^{\phi}_h (a|x) &=   d^{[\pi^{\phi}]^{M}, P, \phi}_h (x) \ls \pi^{\phi} \rs^{M}_h (a|s) =  d^{[\pi^{\phi}]^{M}, P, \phi}_h (x, a),
\end{align*}
where $s \in \phi^{-1}_h (x)$. We finish the proof.    
\end{proof}

\begin{lem} \label{lemma:sample_complexity_of_new_estimator_unknown_transition_state_abstraction}
Given the expert dataset $\gD$, let $\gD$ be divided into two equal subsets, i.e., $\gD = \gD_{1} \cup \gD_{1}^c$ and $\gD_1 \cap \gD_1^{c} = \emptyset$ with $\labs \gD_1 \rabs = \labs \gD_1^{c} \rabs = m / 2$. Let $\pi^{\prime, \phi}$ be the abstract BC's policy on $\gD_1$. Fix $\pi^{\prime, \phi}$, let $\gD^\prime_{\mathrm{env}}$ be the dataset collected by $[\pi^{\prime, \phi}]^M$ and $|\gD^\prime_{\mathrm{env}} | = n^\prime$. Fix $\varepsilon \in (0, 1)$ and $\delta \in (0, 1)$; suppose $H \geq 5$. Consider the abstract state-action distribution estimator $\widetilde{d}^{\piE, \phi}_h$ shown in \eqref{eq:new_estimator_unknown_transition_state_abstraction}, if the expert sample complexity ($m$) and the interaction complexity ($n^\prime$) satisfy
\begin{align*}
    m \gtrsim   \frac{H^{3/2} | \Phi | }{\varepsilon} \log\lp  \frac{|\Phi| H}{\delta} \rp, \; n^\prime \gtrsim \frac{H^{2} | \Phi |}{\varepsilon^2} \log\lp  \frac{|\Phi| H}{\delta} \rp,
\end{align*}
then with probability at least $1-\delta$, we have
\begin{align*}
    \sum_{h=1}^H \lnorm \widetilde{d}^{\piE, \phi}_h - d^{\piE, \phi}_h  \rnorm_{1} \leq \varepsilon.
\end{align*}
\end{lem}

\begin{proof}
First, we can obtain that
\begin{align*}
    \sum_{h=1}^H \lnorm \widetilde{d}^{\piE, \phi}_h - d^{\piE, \phi}_h  \rnorm_1 &= \sum_{h=1}^H \sum_{(x, a) \in \Phi \times \gA} \left\vert \widetilde{d}^{\piE, \phi}_h (x, a) - d^{\piE, \phi}_h (x, a)  \right\vert
    \\
    &=\sum_{h=1}^H \sum_{x \in \Phi} \left\vert \widetilde{d}^{\piE, \phi}_h (x, \pi^{\expert, \phi}_h (x)) - d^{\piE, \phi}_h (x, \pi^{\expert, \phi}_h (x))  \right\vert.
\end{align*}
Here $\pi^{\expert, \phi}$ is the abstract expert policy in \cref{def:abstract_expert_policy}. The last equation holds since $\piE$ is a deterministic policy and satisfies \eqref{eq:expert_consistent} in \cref{asmp:state_abstraction}. Recall the abstract state-action distribution estimator $\widetilde{d}^{\piE, \phi}_h$ shown in \eqref{eq:new_estimator_unknown_transition_state_abstraction}.
\begin{align*}
    \widetilde{d}^{\piE, \phi}_h (x, \pi^{\expert, \phi}_h (x)) &= \frac{\sum_{\tr_h \in \gD^{\prime}_{\env}} \indict \{ \phi_h (\tr_h (\cdot)) = x, \tr_h (a_h) = \pi^{\expert, \phi}_h (x), \tr_h \in \Tr^{\gD_1, \phi}_h   \}}{\vert \gD^{\prime}_{\env} \vert} \nonumber 
    \\
    &+ \frac{\sum_{\tr_h \in \gD_1^c} \indict \{ \phi_h (\tr_h (\cdot)) = x, \tr_h (a_h) = \pi^{\expert, \phi}_h (x), \tr_h \not\in  \Tr^{\gD_1, \phi}_h\}}{\labs \gD^{c}_1 \rabs}.
\end{align*}
Given $\gD_1$, for $d^{\piE, \phi}_h$, we have the following decomposition.
\begin{align*}
    &\quad d^{\piE, \phi}_h (x, \pi^{\expert, \phi}_h (x)) 
    \\
    &= \sum_{\tr_h} \sP^{\piE} (\tr_h) \indict \{ \phi_h (\tr_h (\cdot)) = x, \tr_h (a_h) = \pi^{\expert, \phi}_h (x) \}
    \\
    &= \sum_{\tr_h \in \Tr^{\gD_1, \phi}_h} \sP^{\piE} (\tr_h) \indict \{ \phi_h (\tr_h (\cdot)) = x, \tr_h (a_h) = \pi^{\expert, \phi}_h (x) \} 
    \\
    &+ \sum_{\tr_h \not\in \Tr^{\gD_1, \phi}_h} \sP^{\piE} (\tr_h) \indict \{ \phi_h (\tr_h (\cdot)) = x, \tr_h (a_h) = \pi^{\expert, \phi}_h (x) \} .
\end{align*}
Then we have that
\begin{align*}
    & \quad | \widetilde{d}^{\piE, \phi}_h (x, \pi^{\expert, \phi}_h (x)) - d^{\piE, \phi}_h (x, \pi^{\expert, \phi}_h (x)) |  
    \\
    & \leq \bigg\vert \frac{\sum_{\tr_h \in \gD^{\prime}_{\env}} \indict \{ \phi_h (\tr_h (\cdot)) = x, \tr_h (a_h) = \pi^{\expert, \phi}_h (x), \tr_h \in \Tr^{\gD_1, \phi}_h   \}}{\vert \gD^{\prime}_{\env} \vert}
    \\
    &- \sum_{\tr_h \in \Tr^{\gD_1, \phi}_h} \sP^{\piE} (\tr_h) \indict \{ \phi_h (\tr_h (\cdot)) = x, \tr_h (a_h) = \pi^{\expert, \phi}_h (x) \}   \bigg\vert
    \\
    & + \bigg\vert \frac{\sum_{\tr_h \in \gD_1^c} \indict \{ \phi_h (\tr_h (\cdot)) = x, \tr_h (a_h) = \pi^{\expert, \phi}_h (x), \tr_h \not\in  \Tr^{\gD_1, \phi}_h\}}{\labs \gD^{c}_1 \rabs}
    \\
    &- \sum_{\tr_h \not\in \Tr^{\gD_1, \phi}_h} \sP^{\piE} (\tr_h) \indict \{ \phi_h (\tr_h (\cdot)) = x, \tr_h (a_h) = \pi^{\expert, \phi}_h (x) \}  \bigg\vert.
\end{align*}
We denote the first term in RHS as $\text{EA}_h (x)$ and the second term in RHS as $\text{EB}_h (x)$. We have that
\begin{align*}
    \sum_{h=1}^H \lnorm d^{\piE, \phi}_h  -  \widetilde{d}^{\piE, \phi}_h \rnorm_{1} &\leq \underbrace{\sum_{h=1}^H \sum_{x \in \Phi} \text{EA}_h (x)}_{\text{Error A}} + \underbrace{\sum_{h=1}^H \sum_{x \in \Phi} \text{EB}_h (x)}_{\text{Error B}}.
\end{align*}

First, we analyze the term Error A. Let ${E^\prime}^{x}_h$ be the event that $\tr_h$ agrees with expert policy at abstract state $x$ in time step $h$ and appears in $\Tr_h^{\gD_1, \phi}$. Formally, 
\begin{align*}
    {E^\prime}_h^{x} = \indict\{ \phi_h (\tr_h (\cdot)) = x \cap \tr_h (a_h) = \pi^{\expert, \phi}_h (x) \cap \tr_h \in \Tr_h^{\gD_1, \phi}\}.
\end{align*}

Then we leverage Chernoff's bound to upper bound $\text{EA}_h (x)$. By Lemma \ref{lemma:chernoff_bound}, for each $x \in \gS$ and $h \in [H]$, with probability at least $1 - \frac{\delta}{2 |\Phi| H}$ over the randomness of $\gD ^\prime$, we have
\begin{align*}
   \text{EA}_h (x) \leq \sqrt{ \sP^{\piE} \lp {E^\prime}^{x}_h  \rp  \frac{3 \log \lp 4 |\Phi| H / \delta \rp}{n^\prime}}.
\end{align*}
By union bound, with probability at least $1-\frac{\delta}{2}$ over the randomness of $\gD^\prime_{\env}$, we have
\begin{align*}
    \sum_{h=1}^H \sum_{x \in \Phi} \text{EA}_h (x) &\leq  \sum_{h=1}^H \sum_{x \in \Phi} \sqrt{ \sP^{\piE} \lp {E^\prime}^{x}_h  \rp  \frac{3 \log \lp 4 |\Phi| H / \delta \rp}{n^\prime}}
    \\
    &\leq \sum_{h=1}^H \sqrt{|\Phi|} \sqrt{\sum_{x \in \Phi} \sP^{\piE} \lp {E^\prime}^{x}_h  \rp  \frac{3 \log \lp 4 |\Phi| H / \delta \rp}{n^\prime} }
\end{align*}
The last inequality follows the Cauchy-Schwartz inequality. It remains to upper bound $\sum_{x \in \Phi}  \sP^{\piE}(E_{h}^{x})$ for all $h \in [H]$. To this end, we define the event ${G^\prime}_h^{\gD_1}$ that expert policy $\piE$ visits abstract states covered in $\gD_1$ up to time step $h$. Formally, ${G^\prime}_h^{\gD_1} = \indict\{ \forall h^{\prime} \leq h,  \phi_{h^\prime} (s_{h^{\prime}}) \in \Phi_{h^{\prime}} (\gD_1) \}$, where $\Phi_{h}(\gD_1)$ is the set of abstract states in $\gD_1$ at time step $h$. Then, for all $h \in [H]$, we have \begin{align*}
    \sum_{x \in \Phi} \sP^{\piE} \lp {E^\prime}_h^{x}  \rp = \sP^{\piE}({G^\prime}_h^{\gD_1}) \leq \sP({G^\prime}_1^{\gD_1}).
\end{align*}
The last inequality holds since ${G^\prime}_h^{\gD_1} \subseteq {G^\prime}_1^{\gD_1}$ for all $h \in [H]$. Then we have that
\begin{align*}
    \sum_{h=1}^H \sum_{x \in \Phi} \text{EA}_h (x) \leq H \sqrt{\frac{3 |\Phi| \log \lp 4 |\Phi| H / \delta \rp}{n^\prime}}.
\end{align*}
When the interaction complexity satisfies that $n^\prime \gtrsim \frac{| \Phi | H^{2}}{\varepsilon^2} \log\lp  \frac{|\Phi| H}{\delta} \rp$, with probability at least $1-\frac{\delta}{2}$ over the randomness of $\gD^\prime$, we have $\sum_{h=1}^H \sum_{x \in \Phi} \text{EA}_h (x) \leq \frac{\varepsilon}{2}$.

Second, we upper bound the term Error B. Similarly, we can leverage Chernoff's bound to characterize its concentration rate. For a trajectory $\tr_h$, let $E^x_h$ be the event that $\tr_h$ agrees with expert policy at abstract state $x$ at time step $h$ but is not in $\Tr^{\gD_1, \phi}_h$, that is,
\begin{align*}
    E^x_h = \{ \phi_h (\tr_h (\cdot)) = x \cap \tr_h (a_h) = \pi^{\expert, \phi}_h (x) \cap \tr_h \not\in  \Tr^{\gD_1, \phi}_h \}.
\end{align*}
We consider $E^x_h$ is measured by the stochastic process induced by the expert policy $\piE$. Accordingly, its probability is
denoted as $\sP^{\piE} (E^x_h)$. We see that $\sP^{\piE} (E^x_h)$ is equal to the second term in $\text{EB}_h (x)$. Moreover, the first term in $\text{EB}_h (x)$ is an empirical estimation for $\sP^{\piE} (E^x_h)$. After applying Chernoff's bound, with probability at least $1-\delta/(2 \vert \Phi \vert H)$ with $\delta \in (0, 1)$ (over the randomness of the expert demonstrations $\gD_1^c$), for each $h \in [H], x \in \Phi$, we have
\begin{align*}
    \text{EB}_h (x) \leq \sqrt{\sP ^{\piE} \left(E_{h}^{x}\right) \frac{3 \log (4|\Phi| H / \delta)}{m}}.
\end{align*}
Therefore, with probability at least $1 - \delta/2$, we have
\begin{align*}
     \sum_{h=1}^{H} \sum_{x \in \Phi} \text{EB}_h (x)  &\leq \sum_{h=1}^{H} \sum_{x \in \Phi} \sqrt{\sP ^{\piE} \left(E_{h}^{x}\right) \frac{3 \log (4|\Phi| H / \delta)}{m}}
     \\
     &\leq \sum_{h=1}^{H}  \sqrt{\sum_{x \in \Phi} \sP ^{\piE} \left(E_{h}^{x}\right) \frac{3 \vert \Phi \vert \log (4|\Phi| H / \delta)}{m}},
\end{align*}
where the last step follows the Cauchy–Schwarz inequality. It remains to upper bound $\sum_{x \in \Phi} \sP ^{\piE} \left(E_{h}^{x}\right)$ for all $h \in [H]$. To this end, we define the event $G^{\gD_1}_h$: the expert policy visits certain abstract states uncovered in $\gD_1$ up to time step $h$. Formally, $G^{\gD_1}_h = \{ \exists h^\prime \leq h, \phi_{h^\prime} (s_{h^\prime}) \not \in \Phi_{h^\prime} (\gD_1) \}$, where $\Phi_{h^\prime} (\gD_1)$ is the set of abstract states in $\gD_1$ at time step $h$. Then, for all $h \in [H]$, we have
\begin{align*}
   \sum_{x \in \Phi} \sP ^{\piE} \lp E_{h}^{x} \rp = \sP^{\piE} \lp G^{\gD_1}_h \rp \leq \sP^{\piE} \lp G^{\gD_1}_{H} \rp,  
\end{align*}
where the first equality is true because $\cup_{x \in \Phi} E_{h}^{x} $ corresponds to the event that $\piE$ has visited some state uncovered in $\gD_1$, and the last inequality holds since $G^{\gD_1}_h \subseteq G^{\gD_1}_{H}$ for all $h \in [H]$. Conditioned on $\gD_1$, we further have
\begin{align*}
    \sP(G_H^{\gD_1}) \leq \sum_{h=1}^{H} \sum_{x \in \Phi} d^{\piE, \phi}_h(x) \indict\lb x \notin \Phi_h(\gD_1)  \rb.
\end{align*}

We first consider the expectation $\expect[\sum_{h=1}^{H} \sum_{x \in \Phi} d^{\piE, \phi}_h(x) \indict\lb x \notin \Phi_h(\gD_1)  \rb]$, where the expectation is taken over the expert dataset $\gD_{1}$. 
\begin{align*}
    \expect\ls \sum_{h=1}^{H} \sum_{x \in \Phi} d^{\piE, \phi}_h(x) \indict\lb x \notin \Phi_h(\gD_1)  \rb  \rs \leq \sum_{h=1}^{H} \sum_{x \in \Phi}  d_h^{\piE, \phi}(x) \lp 1 - d_h^{\piE, \phi}(x)  \rp^{m/2} \leq \frac{8 |\Phi| H}{9m},
\end{align*}
where the last step uses the numerical inequality\footnote{The first inequality is based on the basic calculus and the second inequality is based on the fact that $(1 - 1/x)^{x} \leq 1/e \leq 4/9$ while $x \geq 1$.} $\max_{x \in [0, 1]} x (1-x)^{m} \leq {1}/{(1+m)} \cdot \lp 1 - {1}/{m}  \rp^{m} \leq {4}/{(9m)}$. With \citep[Lemma A.3]{rajaraman2020fundamental}, with probability at least $1-\delta$ with $\delta \in (0, \min\{1, H/5\})$, we have
\begin{align*}
    \sum_{h=1}^{H} \sum_{x \in \Phi} d^{\piE, \Phi}_h(x) \indict\lb x \notin \Phi_h(\gD_1)  \rb \leq \frac{8|\Phi| H}{9m} + \frac{6 \sqrt{|\Phi|} H \log(H/\delta)}{m}.
\end{align*}
Then we have 
\begin{align*}
\sum_{h=1}^{H} \sum_{x \in \Phi} \text{EB}_h (x) &\leq \sum_{h=1}^{H} \sqrt{ \lp \frac{8|\Phi| H}{9m} + \frac{6 \sqrt{|\Phi|} H \log(2H/\delta)}{m}  \rp \frac{3 |\Phi| \log (4 |\Phi| H/\delta)}{m}} \\
&\leq \frac{H^{3/2} |\Phi|}{m} \log^{1/2}\lp \frac{4|\Phi| H}{\delta}  \rp \sqrt{ \frac{8}{3} + 18 \log (2H/\delta)  }.
\end{align*}
When the expert sample complexity satisfies that $m \gtrsim   \frac{H^{3/2} | \Phi | }{\varepsilon} \log\lp  \frac{|\Phi| H}{\delta} \rp$, with probability at least $1-\frac{\delta}{2}$ over the randomness of $\gD$, we have $\sum_{h=1}^H \sum_{x \in \Phi} \text{EB}_h (x) \leq \frac{\varepsilon}{2}$. Then, with union bound, with probability at least $1-\delta$, we can obtain
\begin{align*}
    \sum_{h=1}^H \lnorm \widetilde{d}^{\piE, \phi}_h - d^{\piE, \phi}_h  \rnorm_1 \leq \sum_{h=1}^H \sum_{x \in \Phi} \text{EA}_h (x) +  \sum_{h=1}^H \sum_{x \in \Phi} \text{EB}_h (x) \leq \varepsilon,
\end{align*}
which completes the proof.
\end{proof}

\section{Experiment Details}
\label{section:experiment_details}

\textbf{Experiment Setup.} In our experiments, we implement the Reset Cliff MDP with 20 states and 5 actions. The planning horizon is 20. All algorithms are provided with 100 expert trajectories. All experiments run with $20$ random seeds.

\textbf{Algorithm Implementation.} BC directly estimates the expert policy from expert demonstrations. Since the expert policy is deterministic, BC copies the expert action on visited states and takes a uniform policy on non-visited states. The implementation of FEM and GTAL follows the description in \citep{pieter04apprentice} and \citep{syed07game}, respectively.

MB-TAIL first establishes the estimator in \cref{eq:new_estimator_unknown_transition} with $20 \%$ of the environment interactions and learns an empirical transition model by invoking RF-Express~\citep{menard20fast-active-learning} to collect the remaining $80 \%$ trajectories. Subsequently, MB-TAIL performs policy and reward optimization with the recovered transition model. In particular, MB-TAIL utilizes value iteration to obtain the optimal policy (Line 2 of \cref{algo:gradient_based_optimization}). Besides, MB-TAIL utilizes online gradient descent to update the reward function. To utilize the optimization structure, we implement an adaptive step size~\citep{Orabona19a_modern_introduction_to_ol} rather than the constant step size:
\begin{align*}
    \eta_{t} = \frac{D}{ \sqrt{\sum_{i=1}^t \lnorm \nabla_{w} f^{(i)} \lp w^{(i)} \rp \rnorm_2^2}},
\end{align*}
where $D = \sqrt{2H |\gS| |\gA|}$ is the diameter of the set $\gW$. Conclusions about the sample complexity and computational complexity do not change by this adaptive step size. The number of iterations $T$ of MB-TAIL is 500.

To encourage exploration, OAL adds a bonus function to the Q-function. The bonus function used in the theoretical analysis of~\citep{shani2022online} is too big and impractical. Therefore, we simplify their bonus function from $b_{h}^k (s, a)=\sqrt{ \frac{4 H^{2} |\gS| \log \lp 3 H^{2} |\gS| |\gA| n / \delta \rp}{ n_{h}^{k}(s, a) \vee 1}}$ to  $b_{h}^k (s, a)=\sqrt{\frac{ \log \lp H |\gS| |\gA| n / \delta \rp}{n_{h}^{k}(s, a) \vee 1}}$, where $n$ is the total number of interactions, $\delta$ is the failure probability,  $n^{k}_h (s, a)$ is the number of times visiting $(s, a)$ at time step $h$ until episode $k$, and $n^{k}_h(s, a) \vee 1 = \max\{ n^{k}_h(s, a), 1 \}$. With the learned transition model and Q-function, OAL uses mirror descent (MD) to optimize the policy and reward function. The step sizes of MD are set by the results in the theoretical analysis of \citep{shani2022online}. The number of iterations $T$ of OAL is also 500.

\end{document}